%% file: main.tex
\documentclass[11pt]{article}

\usepackage[preprint]{acl}

\usepackage{times}
\usepackage{latexsym}

\usepackage[T1]{fontenc}

\usepackage[utf8]{inputenc}

\usepackage{microtype}

\usepackage{inconsolata}

\usepackage{graphicx}

\usepackage{amsmath,graphicx,hyperref}
\usepackage{amssymb} %
\usepackage{multirow}
\usepackage{booktabs}
\usepackage{subcaption}

\usepackage{enumitem}

\title{Beyond ID Embeddings: Process-Grounded Language Modeling for Cognitive Diagnosis
}

\usepackage{marvosym} %
\usepackage{hyperref} %

\author{
   \textbf{Minghang Liu}$^{\spadesuit,\heartsuit}$,
   \textbf{Yuanzhuo Wang}$^{\spadesuit}$,
   \textbf{Qiang Qiu}$^{\spadesuit}$,
   \\
   \textbf{Huawei Shen}$^{\spadesuit}$,
   \textbf{Xueqi Cheng}$^{\spadesuit}$
   \\
   $^{\spadesuit}$State Key Laboratory of AI Safety, Institute of Computing Technology, CAS\\
   $^{\heartsuit}$University of Chinese Academy of Sciences\\
    {\small
    \texttt{\{liuminghang23s, wangyuanzhuo, qiangqiu, shenhuawei\}@ict.ac.cn}
    }
}

\usepackage[most]{tcolorbox}
\usepackage{xcolor}
\usepackage{tabularx}
\usepackage{booktabs}
\usepackage{array}

\definecolor{LCDInk}{HTML}{1F2937}
\definecolor{LCDBlueBack}{HTML}{F3F7FF}
\definecolor{LCDBlueFrame}{HTML}{4F79B8}
\definecolor{LCDSlateBack}{HTML}{F7F7FA}
\definecolor{LCDSlateFrame}{HTML}{7A8494}
\definecolor{LCDGreenBack}{HTML}{F3FAF5}
\definecolor{LCDGreenFrame}{HTML}{4F9165}
\definecolor{LCDAmberBack}{HTML}{FFF8EE}
\definecolor{LCDAmberFrame}{HTML}{B7791F}

\newtcolorbox{lcdpanel}[4][]{%
  enhanced,
  breakable=false,
  sharp corners=south,
  arc=1.4mm,
  boxrule=0.55pt,
  colback=#2,
  colframe=#3,
  coltitle=LCDInk,
  colbacktitle=#3!18!white,
  fonttitle=\bfseries,
  title={#4},
  left=3.2pt,
  right=3.2pt,
  top=3pt,
  bottom=3pt,
  boxsep=1.5pt,
  before skip=0pt,
  after skip=0pt,
  before upper={\raggedright},
  #1
}

\hypersetup{
  pdftitle={Beyond ID Embeddings: Process-Grounded Language Modeling for Cognitive Diagnosis},
  pdfauthor={Minghang Liu, Yuanzhuo Wang, Qiang Qiu, Huawei Shen, Xueqi Cheng}
}

\begin{document}
\maketitle
\begingroup
\renewcommand{\thefootnote}{\fnsymbol{footnote}}
\endgroup

\input{latex/0_abstract/abstract}

\input{latex/1_introduction/introduction}

\input{latex/2_metod/sum}

\input{latex/3_experiment/sum}

\input{latex/limitations}

\bibliography{custom}

\appendix

\input{latex/appendix/sum}

\end{document}

%% file: latex/0_abstract/abstract.tex
\begin{abstract}

Cognitive Diagnosis Models (CDMs) play a pivotal role in personalized online learning. Traditional CDMs rely on discrete, ID-based embeddings to represent students, exercises, and concepts. This paradigm diverges from the nature of learner cognition, where knowledge is not stored and retrieved as isolated symbols. As a result, CDMs suffer from semantic limitations when new exercises or concepts appear. 
In this paper, we propose a Process-aware Language Cognitive Diagnosis (PLCD) framework that uses language-derived structures as cognitive priors and response records to calibrate student posterior states. PLCD leverages large language models (LLMs) to construct concept schemas and cognitive process graphs, and uses target-conditioned semantic memory to retrieve historical responses that are relevant to each target exercise. 
A process-grounded Language-to-Cognition Mapper with DA-MoE experts and process-level contrastive learning then maps the textual evidence into a unified cognitive space.
Experimental results show that PLCD not only outperforms traditional baselines in predicting student performance but also exhibits strong cognitive transfer capabilities.
These results connect the computational power of LLMs with the psychometric goal of measuring latent knowledge states, suggesting that structured language priors calibrated by response records can improve cold-start robustness and cognitive grounding.

\end{abstract}

%% file: latex/1_introduction/introduction.tex
\section{Introduction}
\begin{figure}[t]          %
    \centering
    \includegraphics[width=\linewidth]{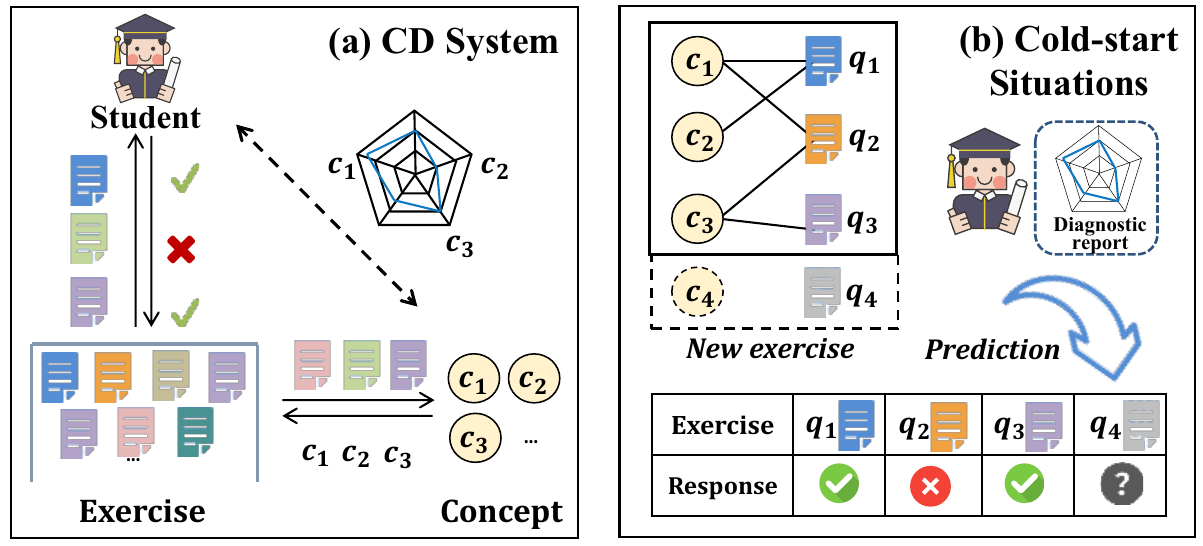}
    \vspace{-1.5em}  %
    \caption{An example of the cognitive diagnosis system and the cold-start problem on traditional ID-based cognitive diagnosis models.}
    \label{fig:motivation}
    \vspace{-1.5em}  %
\end{figure}

In recent years, the evolution of information technology has catalyzed the global growth of online learning platforms and intelligent tutoring systems such as Coursera and ASSISTments \citep{wang2024survey, khajah2014integrating}. Cognitive Diagnosis (CD) stands as a pivotal technique for personalized learner modeling \citep{anderson2014engaging}. As shown in Figure \ref{fig:motivation}(a), a CD system consists of three components: students, exercises, and knowledge concepts. Cognitive Diagnosis Models (CDMs) aim to infer students' mastery levels of specific concepts based on their historical response records.

Although recent self-supervised methods have achieved promising results in CD ~\citep{wang2020neural, li2024towards}, existing methods still exhibit fundamental limitations. 
Traditional CDMs typically construct a tripartite graph consisting of students, exercises, and knowledge concepts ~\citep{ijcai2021p703}. This paradigm places primary reliance on unique ID embeddings to represent each node and models learning as the adjustment of numerical associations between IDs based on response correctness.
However, this contradicts a core tenet of cognitive psychology: human knowledge is structured around meaning, relations, and conceptual similarity \citep{chi1981categorization,anderson2013architecture}. 
ID-centric design offers little access to the compositional semantics of educational content: exercise text, concept descriptions, and their linguistic relations are reduced to arbitrary indices.
As shown in Figure \ref{fig:motivation}(b), when a new exercise appears, an ID-based CDM has no direct way to utilize its textual and logical content unless sufficient response records are observed. This raises a fundamental issue: ID-based CDMs fail to capture the underlying structure of students' cognitive states and instead mainly fit behavioral regularities.

Recent work has begun to incorporate textual and external knowledge into CDMs. Some studies use knowledge graphs to improve cold-start transfer \citep{gao2023leveraging}, while others employ LLMs or text encoders to enrich CDMs with semantic features \citep{Dong_Chen_Wu_2025, zhao2025multi}. 
These efforts show that language provides useful educational signals beyond sparse response records. 
However, most existing approaches still treat language as an auxiliary enhancement to ID-based CDMs: student and exercise IDs remain the primary representational anchors, while textual semantics mainly serve as side information for representation alignment. 
They also tend to compress a learner’s entire history into a fixed student profile, potentially obscuring target-relevant cognitive evidence and dynamic mastery patterns. 
As a result, these methods still struggle to transcend the inherent limitations of ID embeddings.
We discuss broader related work on ID-based CDMs and language-enhanced diagnosis in Appendix~\ref{app:related_work}.

In this study, we propose the Process-aware Language Cognitive Diagnosis (PLCD) framework that uses language-derived structures as cognitive priors and learner response records for personalized calibration. 
Unlike prior methods that treat text as auxiliary features for ID embeddings, PLCD models educational content as structured cognitive evidence without relying on latent vectors tied to arbitrary student IDs. 
It first employs LLMs to construct concept schemas and cognitive process graphs, then uses target-conditioned semantic memory to retrieve cognitively similar historical responses for prediction. 
Through Language-to-Cognition Mapping, a DA-MoE transforms semantic evidence into diagnostic representations. 
Finally, process-level supervised contrastive learning sharpens the semantic distinction between mastered and unmastered concepts, while a psychometric outcome head accounts for exercise-level guessing and slipping effects to infer learning outcomes.

Our experiments across three real-world educational datasets show that PLCD can reliably infer learning outcomes from inferred cognitive states and significantly outperform ID-based models.
These findings suggest that structured language priors, when calibrated by response records, can improve cold-start robustness and provide a cognitively grounded alternative to ID-centric representation learning.
Furthermore, qualitative analysis shows that PLCD can distinguish different types of cognitive deficits, such as semantic translation errors and procedural failures, thus offering a new computational lens for studying how knowledge is represented, retrieved, and applied in educational tasks.

%% file: latex/2_metod/sum.tex
\section{Methodology}

\input{latex/2_metod/2.1}
\input{latex/2_metod/2.2}
\input{latex/2_metod/2.3}

%% file: latex/2_metod/2.1.tex
\subsection{Structured Cognitive Evidence Construction}
~\label{2.1}

\begin{figure*}[t]
    \centering
    \includegraphics[width=1\linewidth]{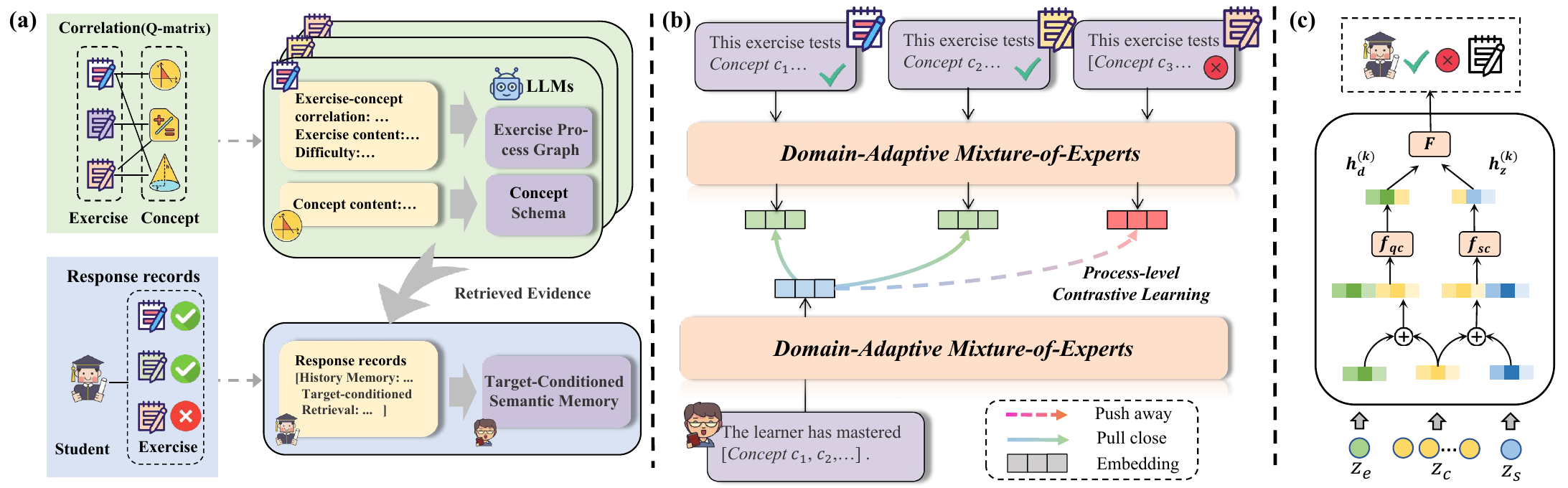}
    \vspace{-1.5em}
    \caption{The overall framework of Process-aware Language Cognitive Diagnosis (PLCD). (a) Structured Cognitive Evidence Construction. (b) Language-to-Cognition Mapping for Diagnosis. (c) Inferring Learning Outcomes from Cognitive States.
 }
    \label{fig:model}
  \vspace{-1.0em} 
\end{figure*}

\textbf{Concept Schemas and Cognitive Process Graphs}. PLCD first converts textual educational content into structured cognitive evidence rather than free-form semantic profiles. Large language models (LLMs) serve as cognitive priors: before any learner-specific calibration, they externalize concept definitions, prerequisite relations, subskills, common misconceptions, and the operations needed to solve an exercise.
For each concept $c$, we construct a typed concept schema:
\begin{equation}
\mathcal{G}_c = \operatorname{LLM}(M_c, I_c), \quad I_c = \operatorname{name}(c),
\end{equation}
where $M_c$ prompts the LLM to populate fields including definition, prerequisites, subskills, common misconceptions, and related cognitive operations. This schema represents the expert-side cognitive structure associated with a knowledge concept.

For each exercise $e$, PLCD constructs a cognitive process graph $\mathcal{G}_e$ instead of a natural-language exercise profile:
\begin{equation}
\small
\mathcal{G}_e = \operatorname{LLM}(M_e, I_e), \quad
I_e = [d_e, \{c \mid Q_{e,c}=1\}, l_e],
\end{equation}
where $d_e$ is the exercise text, $\{c \mid Q_{e,c}=1\}$ is the concept set indicated by the Q-matrix, and $l_e$ is the difficulty level. The resulting graph is defined as:
\begin{equation}
\mathcal{G}_e =
\langle \mathcal{C}_e, \mathcal{O}_e, \mathcal{U}_e, \mathcal{S}_e, \mathcal{D}_e \rangle .
\end{equation}
Here, $\mathcal{C}_e=\{(c,w_c)\}$ contains required concepts with importance weights, $\mathcal{O}_e=\{(o,w_o)\}$ contains cognitive operations such as semantic-to-symbolic translation and procedural manipulation, $\mathcal{U}_e$ stores textual cues that trigger these operations, $\mathcal{S}_e$ records solution steps, and $\mathcal{D}_e$ lists potential error types or misconceptions. This makes exercise evidence explicitly modelable by moving beyond coarse concept membership to capture the reasoning operations required by each exercise and the error-prone steps in its solution process.

\textbf{Target-Conditioned Semantic Memory}.PLCD does not generate a static student profile. Compressing all historical responses into one global paragraph can obscure how a learner behaved on tasks that are cognitively similar to the target exercise. Instead, for each student $s$ at time $t$, PLCD preserves a history memory:
\begin{equation}
\mathcal{M}_{s,t} = \{(\mathcal{G}_{e_i}, y_i, \Delta t_i) \mid i < t\},
\end{equation}
where $\mathcal{G}_{e_i}$ is the cognitive process graph of a previously attempted exercise, $y_i$ is the correctness label, and $\Delta t_i$ is the temporal interval between that attempt and the target prediction time.

When predicting student $s$ on a target exercise $e$, PLCD retrieves evidence conditioned on the target graph $\mathcal{G}_e$. Let $\mathbf{h}_e$ and $\mathbf{h}_{e_i}$ be graph-level embeddings of the target and historical exercises, and let $\mathbf{q}_e$ and $\mathbf{q}_{e_i}$ be their concept-incidence vectors derived from the Q-matrix. The relevance of a historical exercise is computed as:
\begin{equation}
\small
\alpha_i =
\frac{
\exp(\mathbf{h}_e^\top \mathbf{h}_{e_i}/\tau_m + \gamma \mathbf{q}_e^\top \mathbf{q}_{e_i})
}{
\sum_{j<t}\exp(\mathbf{h}_e^\top \mathbf{h}_{e_j}/\tau_m + \gamma \mathbf{q}_e^\top \mathbf{q}_{e_j})
},
\end{equation}
where $\tau_m$ controls the sharpness of semantic retrieval and $\gamma$ balances graph-level semantic similarity with concept overlap. The target-conditioned learner evidence is then:
\begin{equation}
\mathbf{m}_{s,e,t} = \sum_{i<t} \alpha_i \cdot \psi(\mathcal{G}_{e_i}, y_i, \Delta t_i),
\end{equation}
where $\psi(\cdot)$ encodes a historical cognitive graph together with correctness and temporal information. The vector $\mathbf{m}_{s,e,t}$ is the student evidence representation for the current target: it asks which prior exercises required similar cognitive operations, whether the student solved them correctly, and which process-level errors may be implicated.

Finally, we employ a pre-trained language model $\mathcal{T}$ as a universal text embedder by serializing structured graphs into text:
\begin{equation}
\small
\mathbf{z}_{s \mid e,t}=\mathbf{m}_{s,e,t},\;
\mathbf{z}_e=\mathcal{T}(\operatorname{ser}(\mathcal{G}_e)),\;
\mathbf{z}_c=\mathcal{T}(\operatorname{ser}(\mathcal{G}_c)),
\end{equation}
where $\mathbf{z}_{s \mid e,t}$ is a target-conditioned student representation and $\mathbf{z}_e,\mathbf{z}_c \in \mathbb{R}^{d}$ are exercise and concept representations.

It is important to clarify that PLCD is not response-free. Language provides a zero-shot cognitive prior, while response records calibrate the learner-specific posterior state through target-conditioned memory. IDs are used only as retrieval indices during training, which is unavoidable in supervised educational data. They are never learned as embeddings, optimized as parameters, or used as input features. 

%% file: latex/2_metod/2.2.tex
\subsection{Language-to-Cognition Mapping for Diagnosis}

This section describes how PLCD maps language-derived cognitive evidence into diagnostic states. Instead of treating language space as a complete substitute for response data, PLCD uses language-derived structures as cognitive priors and calibrates them with learner response records. The mapping stage has two components: a process-grounded DA-MoE mapper and a process-level supervised contrastive objective.

\textbf{Process-Grounded DA-MoE Mapper}. Educational data exhibits high cognitive heterogeneity: for example, geometric reasoning, algebraic manipulation, and proportional reasoning require different mental operations. A standard MoE layer can model such heterogeneity, but abstract neural experts are hard to interpret cognitively. We therefore bind each expert to a predefined cognitive process type extracted from the cognitive process graphs in Section~\ref{2.1}. The full expert-process inventory is provided in Appendix~\ref{app:process_expert_inventory}.

Let $\mathcal{P}=\{p_1,\ldots,p_N\}$ denote the process inventory. For an exercise $e$, the cognitive process graph $\mathcal{G}_e$ provides weighted operations $\mathcal{O}_e=\{(o,w_o)\}$. We convert them into a process prior:
\begin{equation}
\small
\pi_{e,p} =
\frac{\sum_{(o,w_o)\in \mathcal{O}_e} w_o \mathbb{I}[\eta(o)=p]}
{\sum_{p'\in\mathcal{P}}\sum_{(o,w_o)\in \mathcal{O}_e} w_o \mathbb{I}[\eta(o)=p'] + \varepsilon},
\end{equation}
where $\eta(\cdot)$ maps an LLM-generated operation label to the closest process type in $\mathcal{P}$, and $\varepsilon$ avoids division by zero. We write $\boldsymbol{\pi}_e=[\pi_{e,p}]_{p\in\mathcal{P}}$ for the full process-prior vector.

We construct a set of process experts $\{E_p\}_{p\in\mathcal{P}}$, where each expert is a feed-forward network. Given an input representation $z$--which can be a student representation $\mathbf{z}_{s \mid e,t}$, an exercise representation $\mathbf{z}_e$, or a concept representation $\mathbf{z}_c$, the output of expert is:
\begin{equation}
\small
r_p = E_p(z) = \operatorname{ReLU}(W_{p,2}\operatorname{ReLU}(W_{p,1}z+b_{p,1})+b_{p,2}).
\end{equation}

The gate is conditioned on both the input embedding and the target exercise's process prior:
\begin{equation}
\small
\mathbf{g}(z,e)=\operatorname{Softmax}\left(\operatorname{TopK}\left(W_gz+\lambda_g\log(\boldsymbol{\pi}_e+\varepsilon), k\right)\right),
\end{equation}
\begin{equation}
\small
\hat{\mathbf{z}}_{\mid e}=\sum_{p\in\mathcal{P}}[\mathbf{g}(z,e)]_p\cdot r_p .
\end{equation}
Here, $\lambda_g$ controls the strength of the cognitive process prior, and $\operatorname{TopK}(\cdot,k)$ activates only the $k$ most relevant process experts. Applying this mapper to $\mathbf{z}_{s \mid e,t}$, $\mathbf{z}_e$, and $\mathbf{z}_c$ yields $\hat{\mathbf{z}}_{s \mid e,t}$, $\hat{\mathbf{z}}_e$, and $\hat{\mathbf{z}}_c$, respectively. This gate preserves neural flexibility while making the activated experts traceable to explicit reasoning operations. For exercise representations, we further regularize the gate toward the process prior:
\begin{equation}
\mathcal{L}_{proc}=\sum_{e\in\mathcal{E}}\operatorname{KL}\left(\boldsymbol{\pi}_e \,\|\, \mathbf{g}(\mathbf{z}_e,e)\right).
\end{equation}

To prevent the collapse problem where the gating network favors only a few experts, we retain an auxiliary load balancing loss:
\begin{equation}
\mathcal{L}_{aux} = N \sum_{p\in\mathcal{P}} f_p \cdot P_p ,
\end{equation}
where $f_p$ is the fraction of samples assigned to expert $p$, and $P_p$ is the average gating probability for expert $p$. Combining the above two loss terms, the overall loss for MoE framework is defined as:
\begin{equation}
\mathcal{L}_{MoE}=\mathcal{L}_{proc}+\beta\mathcal{L}_{aux}.
\end{equation}

\textbf{Process-Level Supervised Contrastive Learning}. A coarse exercise-level contrastive objective would treat a correctly answered exercise as a positive sample and an incorrectly answered exercise as a negative sample. This is too coarse for cognitive diagnosis: an exercise may require multiple concepts and processes, and a wrong answer does not imply that the student failed every required process. We therefore perform contrastive learning at the process level.

For each process $p$, we compute the student's historical support and error evidence under the target-conditioned retrieval weights from Section~2.1:
\begin{equation}
u_{s,e,t,p}=\sum_{i<t}\alpha_i y_i \pi_{e_i,p},
v_{s,e,t,p}=\sum_{i<t}\alpha_i (1-y_i) \pi_{e_i,p}.
\end{equation}
Here, $u_{s,e,t,p}$ summarizes prior success on similar process demands, while $v_{s,e,t,p}$ summarizes prior errors. For a target interaction $(s,e,t,y_{s,e,t})$, we define positive and negative weights:
\begin{equation}
\small
\begin{aligned}
a^+_{s,e,t,p}
&= \pi_{e,p}\left[
y_{s,e,t}
+(1-y_{s,e,t})
\frac{u_{s,e,t,p}}{u_{s,e,t,p}+v_{s,e,t,p}+\varepsilon}
\right], \\
a^-_{s,e,t,p}
&= \pi_{e,p}(1-y_{s,e,t})
\frac{v_{s,e,t,p}}{u_{s,e,t,p}+v_{s,e,t,p}+\varepsilon}.
\end{aligned}
\end{equation}

Thus, correct responses provide positive evidence for the required processes, while incorrect responses produce negative evidence only for processes that are both required by the target exercise and historically weak for the student.

The process-specific matching score is computed before expert aggregation:
\begin{equation}
\operatorname{score}_p(s,e,t)=\operatorname{sim}\left(E_p(\mathbf{z}_{s \mid e,t}), E_p(\mathbf{z}_e)\right),
\end{equation}
where $\operatorname{sim}(\cdot,\cdot)$ denotes cosine similarity. The process-level contrastive loss is:
\begin{equation}
\small
\begin{aligned}
B_{s,p}
&=
\sum_{(e_k,t_k)\in\mathcal{N}_{s,p}}
a^-_{s,e_k,t_k,p}
\exp(\operatorname{score}_p(s,e_k,t_k)/\tau), \\
\ell_{s,e,t,p}
&=
-a^+_{s,e,t,p}
\log
\frac{\exp(\operatorname{score}_p(s,e,t)/\tau)}
{\exp(\operatorname{score}_p(s,e,t)/\tau)+B_{s,p}}, \\
\mathcal{L}_{PCL}
&=
\sum_{(s,e,t)\in\mathcal{R}}
\sum_{p\in\mathcal{P}}
\ell_{s,e,t,p}.
\end{aligned}
\end{equation}

where $\mathcal{N}_{s,p}$ is the set of sampled interactions for student $s$ with nonzero negative evidence on process $p$, and $\tau$ is a temperature hyperparameter. This objective aligns the learner representation with the specific cognitive operations that the evidence supports, rather than pushing whole exercises together or apart solely by correctness labels.

%% file: latex/2_metod/2.3.tex
\vspace{-0.7em}
\subsection{Inferring Learning Outcomes from Cognitive States}

For each response record $(s,e,t,y_{s,e,t})$, the target exercise $e$ contains a concept set $\mathcal{C}_e=\{c_k\}_{k=1}^{n}$. 
For notational simplicity, in this subsection we use $\hat{\mathbf{z}}_{s}$ as a shorthand for $\hat{\mathbf{z}}_{s \mid e,t}$, since each prediction is made for a fixed response record $(s,e,t,y_{s,e,t})$.
The model generates concept-specific student knowledge states \(\textbf{\textit{h}}_z^{(k)}\) and exercise difficulty factors \(\textbf{\textit{h}}_d^{(k)}\), which are used to infer the probability that students answer the exercise correctly:
\begin{equation}
\begin{aligned}
\mathbf{h}_{z}^{(k)}
&= \sigma\!\left(f_{sc}\!\left(\hat{\mathbf{z}}_{s } \oplus \hat{\mathbf{z}}_{c_k}\right)\right), \\
\mathbf{h}_{d}^{(k)}
&= \sigma\!\left(f_{qc}\!\left(\hat{\mathbf{z}}_{e} \oplus \hat{\mathbf{z}}_{c_k}\right)\right), \\
r_{s,e,t}
&= \frac{1}{|\mathcal{C}_e|}
\sum_{c_k \in \mathcal{C}_e}
\sigma\!\left(F\!\left(\mathbf{h}_{z}^{(k)} - \mathbf{h}_{d}^{(k)}\right)\right).
\end{aligned}
\end{equation}
Here, \(\oplus\) denotes vector concatenation, $f_{sc}$ and $f_{qc}$ are neural networks, and $F(\cdot)$ is a fully connected layer. The intermediate score $r_{s,e,t}$ represents the diagnosis-based probability before accounting for response noise. 

Inspired by item response theory and the slip-guess parameterization in cognitive diagnosis models \citep{junker2001cognitive,delatorre2009dina}, we further augment this diagnostic score with exercise-level guessing and slipping effects:
\begin{equation}
\small
[g_e,\ell_e,\rho_e]=\operatorname{Softmax}\left([\mathbf{w}_g^\top\hat{\mathbf{z}}_e+b_g,\mathbf{w}_{\ell}^\top\hat{\mathbf{z}}_e+b_{\ell},0]\right),
\end{equation}
where $g_e$ is the probability of guessing correctly, $\ell_e$ is the probability of slipping despite sufficient mastery, and the residual term $\rho_e$ ensures $g_e,\ell_e\in[0,1]$ and $g_e+\ell_e<1$. The final response probability is:
\begin{equation}
\hat{y}_{s,e,t}=g_e+(1-g_e-\ell_e)r_{s,e,t},
\end{equation}
which preserves the original diagnostic structure while adding a psychometric correction for exercise-level response noise.

The response prediction loss is the negative log-likelihood of observed correctness labels:
\begin{equation}
\mathcal{L}_{cd} = -\sum_{(s,e,t,y) \in \mathcal{R}} [y \log(\hat{y}) + (1 - y) \log(1 - \hat{y})],
\end{equation}
where $\mathcal{R}$ is the set of all student response records.
The total loss is defined as:
\begin{equation}
\mathcal{L} = \mathcal{L}_{cd} + \lambda_{1}\mathcal{L}_{PCL} + \lambda_{2}\mathcal{L}_{\mathrm{MoE}},
\end{equation}
where $\lambda_{1}$ and $\lambda_{2}$ control the contribution of these two auxiliary terms. This objective keeps supervised response calibration explicit while preserving the zero-shot cognitive prior supplied by language-derived process graphs.

%% file: latex/3_experiment/sum.tex
\section{Experiments}

\input{latex/3_experiment/experimental_setup}

\input{latex/3_experiment/2}

\input{latex/3_experiment/3}

\input{latex/3_experiment/6}

\input{latex/3_experiment/7}

\input{latex/3_experiment/Visualization}

\input{latex/3_experiment/5}

\section{Conclusion}

In this paper, we propose \textbf{PLCD}, a process-aware language cognitive diagnosis framework that avoids learned student and exercise ID embeddings by using structured language-derived cognitive priors.
Our experiments show that these priors, when calibrated by response records through target-conditioned memory, improve diagnostic accuracy and robustness in sparse and cold-start settings.
Qualitative analyses further show that PLCD captures psychologically meaningful structure in latent space and can distinguish different types of cognitive deficits, such as semantic translation errors and procedural failures. Taken together, our findings suggest that structured educational language, paired with explicit response calibration, provides a promising foundation for cognitively grounded approaches to cognitive diagnosis.

%% file: latex/3_experiment/experimental_setup.tex
\subsection{Experimental Setup}

We evaluate PLCD on three real-world intelligent education datasets, Junyi~\citep{chang2015modeling}, XES3G5M~\citep{liu2023xes3g5m}, and MOOC~\citep{yu2023moocradar}, and compare it with representative ID-based and knowledge-enhanced cognitive diagnosis models. We report ACC, AUC, and RMSE for student performance prediction, and further use Brier Score and Expected Calibration Error (ECE) in component-level calibration analysis. Detailed dataset statistics, baseline descriptions, implementation settings, and hyperparameter choices are provided in Appendix~\ref{app:experimental_basic_setup}.

%% file: latex/3_experiment/2.tex
\vspace{-0.5em}
\subsection{Performance Comparison and Ablation Study}

\begin{table*}[t]
  \centering
  \resizebox{\textwidth}{!}{
  \begin{tabular}{@{}lccccccccc@{}}
    \toprule
    \multirow{2}{*}{\textbf{Model}} 
    & \multicolumn{3}{c}{\textbf{Junyi}} 
    & \multicolumn{3}{c}{\textbf{XES3G5M}} 
    & \multicolumn{3}{c}{\textbf{MOOC}} \\
    \cmidrule(lr){2-4} \cmidrule(lr){5-7} \cmidrule(lr){8-10}
    & ACC\%$\uparrow$ & AUC\%$\uparrow$ & RMSE\%$\downarrow$
    & ACC\%$\uparrow$ & AUC\%$\uparrow$ & RMSE\%$\downarrow$
    & ACC\%$\uparrow$ & AUC\%$\uparrow$ & RMSE\%$\downarrow$ \\
    \midrule
    IRT  
    & 67.60 & 77.50 & 42.68 
    & 72.90 & 76.05 & 41.92 
    & 73.23 & 73.30 & 41.85 \\
    MIRT 
    & 74.70 & 79.16 & 41.17 
    & 73.23 & 76.55 & 41.66 
    & 72.11 & 71.31 & 42.01 \\
    \midrule
    NCD  
    & 74.43 & 79.09 & 41.72 
    & 74.60 & 78.04 & 41.12 
    & 78.64 & 79.31 & 40.12 \\
    RCD  
    & 76.33 & 81.57 & 40.41 
    & 77.15 & 81.30 & 40.17 
    & 77.01 & 77.51 & 40.62 \\
    ACD  
    & 76.80 & 81.80 & 40.59 
    & 75.10 & 79.66 & 41.53 
    & 79.23 & 79.81 & 39.53 \\
    \midrule
    KaNCD  
    & 75.60 & 79.33 & 41.05 
    & 78.89 & 82.19 & 39.56 
    & 77.22 & 78.01 & 40.55 \\
    TechCD 
    & 76.10 & 81.22 & 40.83 
    & 78.80 & 82.18 & 40.03 
    & 76.01 & 76.98 & 40.78 \\
    KCD    
    & \underline{78.03} & \underline{82.29} & \underline{40.21} 
    & \underline{80.01} & \underline{82.97} & \underline{39.04} 
    & 80.22 & 81.01 & 39.27 \\
    DMC-CDM 
    & 77.50 & 82.02 & 40.49  
    & 79.67 & 82.70 & 39.13 
    & \underline{83.81} & \underline{84.31} & \underline{38.91} \\
    \midrule
    \textbf{PLCD (Ours)} 
    & \textbf{80.81}$^{*}$ & \textbf{84.52}$^{*}$ & \textbf{39.01}$^{*}$ 
    & \textbf{83.51}$^{*}$ & \textbf{85.37}$^{*}$ & \textbf{38.13}$^{*}$ 
    & \textbf{87.16}$^{*}$ & \textbf{88.33}$^{*}$ & \textbf{36.51}$^{*}$ \\
    \quad \textit{w/o Structured Evidence} 
    & 76.52 & 81.67 & 40.51 
    & 77.29 & 81.94 & 40.11 
    & 78.95 & 80.21 & 39.45 \\
    \quad \textit{w/o Target-conditioned Memory}
    & 78.73 & 82.88 & 39.84
    & 79.96 & 82.57 & 39.81
    & 80.02 & 81.69 & 39.35 \\
    \quad \textit{w/o Process-grounded Mapper} 
    & 78.51 & 82.80 & 39.92 
    & 80.36 & 83.12 & 39.25
    & 82.74 & 83.85 & 38.52 \\
    \quad \textit{w/o Process-level CL}
    & 79.31 & 83.36 & 39.44
    & 82.09 & 83.74 & 38.86
    & 85.47 & 86.12 & 37.48 \\
    \bottomrule
  \end{tabular}
  }
  \caption{Experimental results and ablation study on student performance prediction. The most competitive baseline results are \underline{underlined}, and the best results are highlighted in \textbf{bold}. The superscript $^{*}$ denotes a statistically significant improvement over the strongest baseline under a two-sided paired $t$-test across the same ten runs ($p<0.05$).}
  \label{tab:main}
  \vspace{-1.2em}
\end{table*}

Table~\ref{tab:main} compares PLCD with a range of state-of-the-art models, showing that PLCD consistently outperforms all baselines. This result supports our claim that structured language-derived evidence provides a useful cognitive prior, while response records calibrate the learner-specific posterior state through target-conditioned memory. Crucially, PLCD shows significant advantages even when compared to recent knowledge-enhanced CDMs (e.g., KCD, DMC-CDM) that incorporate external knowledge as auxiliary features to ID-based frameworks. This suggests that language should not merely serve as side information for ID embeddings, but can organize the diagnostic representation itself when paired with explicit response calibration.

To validate the effectiveness of the major components, we report four ablation variants:
(1) \textit{w/o Structured Evidence}, where we remove concept schemas and cognitive process graphs and instead encode raw exercise text and concept names;
(2) \textit{w/o Target-conditioned Memory}, where target-conditioned retrieval is removed, and historical responses are aggregated independent of target exercises; 
(3) \textit{w/o Process-grounded Mapper}, where we remove the DA-MoE mapper and rely on the initial language embeddings; and 
(4) \textit{w/o Process-level CL}, where the process-level supervised contrastive objective is removed while the prediction objective is retained.

As shown in Table~\ref{tab:main}, the performance of \textit{w/o Structured Evidence} drops across all datasets, indicating that raw text alone is insufficient to expose prerequisite relations, process-level operations, textual cues, and likely misconception types. 
The \textit{w/o Target-conditioned Memory} variant also yields consistent degradation, especially on XES3G5M and MOOC, where students have richer response histories. This suggests that aggregating all historical responses into a static learner profile blurs the evidence most relevant to the target exercise, while target-conditioned retrieval better captures whether the learner succeeded or failed on cognitively similar tasks. 
The \textit{w/o Process-grounded Mapper} variant further leads to a noticeable drop, showing that pre-trained language embeddings are not automatically aligned with diagnostic states and must be transformed into cognitively meaningful representations. 
Finally, removing process-level contrastive learning causes a smaller but stable decline, confirming that fine-grained process supervision improves the discriminability of learner states beyond coarse exercise-level correctness signals.

%% file: latex/3_experiment/3.tex
\subsection{Cold-start and Sparsity Analysis
}

\begin{figure}[t]          %
    \centering
    \includegraphics[width=\linewidth]{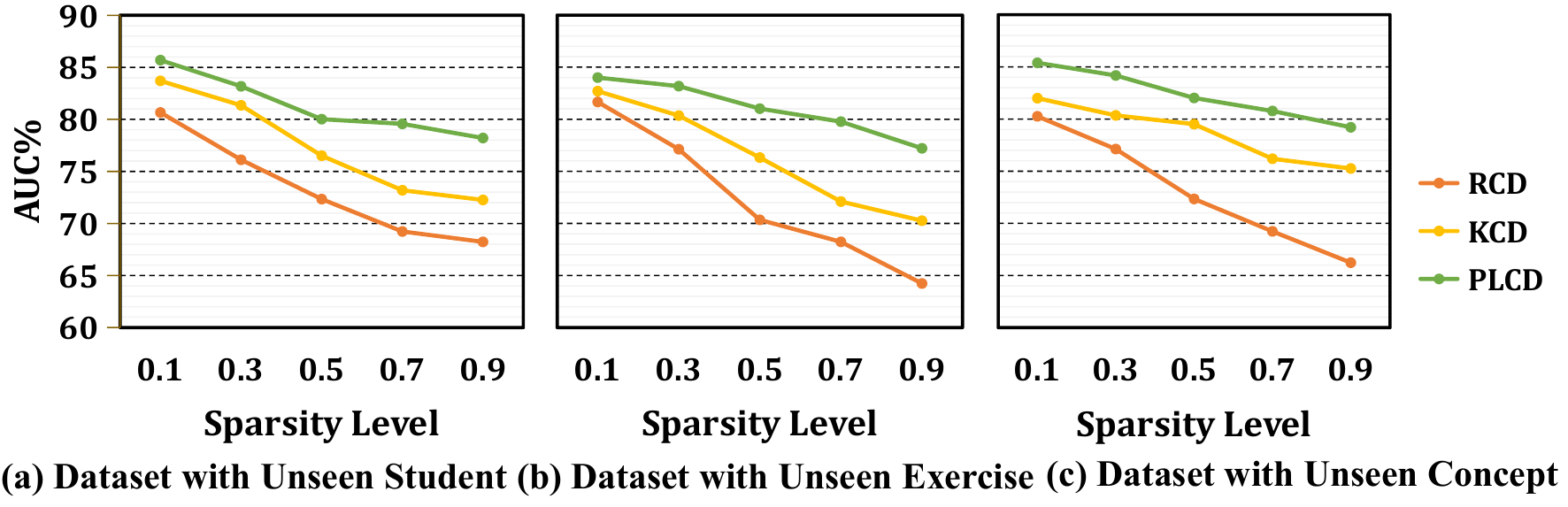}
    \vspace{-1.5em}
    \caption{Comparison with CDMs in different sparsity levels.}
    \label{fig:cold}
    \vspace{-1.0em}
\end{figure}

\begin{table}[t]
  \centering
  \small
  \resizebox{1.0\linewidth}{!}{ %
    \begin{tabular}{llcccc}
        \toprule
        \multirow{2.5}{*}{\textbf{Cold-start}} &
        \multirow{2.5}{*}{\textbf{Metrics}} &
        \multicolumn{4}{c}{\textbf{Models}} \\
        \cmidrule(lr){3-6} 
        & & RCD & KCD & PLCD & \textit{Oracle} \\
        \midrule
        \multirow{2}{*}{New Concepts} 
            & ACC\% & 69.27 & 70.98 & 75.58 & \textit{83.51} \\
            & AUC\% & 68.93 & 72.72 & 77.92 & \textit{85.37} \\
        \midrule
        \multirow{2}{*}{New Exercises} 
            & ACC\% & 66.46 & 69.09 & 73.09 & \textit{83.51} \\
            & AUC\% & 63.78 & 69.55 & 73.19 & \textit{85.37} \\
        \midrule
        \multirow{2}{*}{Missing Q-matrix} 
            & ACC\% & -- & 75.94 & 80.46 & \textit{83.51} \\
            & AUC\% & -- & 79.00 & 82.37 & \textit{85.37} \\
        \bottomrule
    \end{tabular}
    }
    \caption{Performance on cold-start prediction for new concepts, new exercises, and missing Q-matrix settings on XES3G5M. \textit{Oracle} denotes the full-test PLCD result and is included only as a reference upper bound.}
    \vspace{-1.5em}
    \label{tab:cold}
\end{table}

A primary challenge for ID-based CDMs is their performance degradation in data-scarce and cold-start situations. To assess the robustness of our model, we first evaluate its performance under varying levels of data sparsity. 
As shown in Figure~\ref{fig:cold}, PLCD consistently outperforms both the traditional ID-based model RCD and the knowledge-enhanced model KCD across all tested sparsity levels on XES3G5M. In particular, the performance advantage of PLCD becomes more pronounced as response records become sparse. This behavior is consistent with our design: language-derived process graphs provide reusable cognitive priors, while target-conditioned memory uses the few available historical responses to calibrate learner-specific evidence for the current target exercise.
We further investigate three cold-start settings, namely new concepts, new exercises, and missing Q-matrix. Detailed experimental settings are presented in Appendix~\ref{app:cold_start_protocol}.
Table~\ref{tab:cold} shows that PLCD consistently outperforms the baselines across the three cold-start and annotation-missing settings. 
For new concepts, PLCD exceeds KCD by 4.60 ACC points, suggesting that language-derived cognitive structures support transfer to concepts without interaction history. 
For new exercises, PLCD improves over KCD by 4.00 ACC points, showing that exercise text and process graphs provide useful item-side evidence when learned item embeddings are unavailable. 
In the missing-Q-matrix setting, PLCD still exceeds KCD by 4.52 ACC points, indicating that exercise text and inferred process structures can partially compensate for fully masked positive exercise--concept links. 
The remaining gap to Oracle confirms that explicit Q-matrix annotations remain useful, but PLCD is less dependent on them than Q-matrix-centered baselines.

%% file: latex/3_experiment/6.tex
\subsection{Mechanism Verification}

We further verify whether the proposed target-conditioned semantic memory retrieves behaviorally meaningful learner evidence. To isolate the effect of the memory mechanism, we replace only the retrieval weights used to construct the learner memory vector. The detailed experimental setup is provided in Appendix~\ref{app:memory_setup}.

\begin{table}[t]
\centering
\resizebox{\columnwidth}{!}{
\begin{tabular}{lccc}
\toprule
\multirow{2}{*}{Memory Strategy} 
& Prediction 
& \multicolumn{2}{c}{Retrieval-level} \\
\cmidrule(lr){2-2} \cmidrule(lr){3-4}
& ACC\% 
& Q-Jac.@5$\uparrow$& Res.Corr.@5$\uparrow$ \\
\midrule
Static Learner Summary 
& 79.96 & -- & -- \\
Random Retrieval 
& 78.72 & 0.118 & 0.018 \\
Concept-overlap Only 
& 80.47 & 0.386 & 0.086 \\
Semantic Only 
& 81.36 & 0.274 & 0.112 \\
Semantic w/ Shuffled Labels 
& 80.21 & \textbf{0.412} & 0.009 \\
Semantic+Q Retrieval (PLCD) 
& \textbf{83.51} & \textbf{0.412} & \textbf{0.168} \\
\bottomrule
\end{tabular}
}
\caption{
Mechanism verification of target-conditioned semantic memory. 
Q-Jac.@5 and Res.Corr.@5 are retrieval-level diagnostics defined in Appendix~\ref{app:retrieval_diagnostics}.
}
\label{tab:memory_mechanism}
\end{table}

The results show that our method achieves the best prediction performance. 
Compared with semantic-only and concept-overlap retrieval, PLCD obtains higher ACC, indicating that graph-level semantic similarity and Q-matrix concept structure provide complementary retrieval signals. 
Random retrieval performs the worst among the retrieval-based variants, confirming that the gain does not come from simply adding arbitrary historical responses.
The label-shuffled variant further verifies the behavioral role of memory. 
Although it retrieves the same historical exercises as Semantic+Q retrieval, its ACC drops markedly, showing that PLCD benefits from the student's actual correctness patterns on the retrieved exercises rather than exercise similarity alone. 
The retrieval diagnostics support this interpretation: Semantic+Q retrieval achieves the highest Q-Jac.@5, while its Res.Corr.@5 is much higher than the label-shuffled variant. This suggests that the retrieved correctness labels provide learner-specific predictive evidence beyond conceptual relevance.

%% file: latex/3_experiment/7.tex
\subsection{Language-to-Cognition Mapping Analysis}

\begin{table}[t]
  \centering
  \resizebox{1.0\linewidth}{!}{
  \begin{tabular}{lcccc}
    \toprule
    \textbf{Model Variant} & ACC\% & GP-Cos$\uparrow$ & KL$\downarrow$ & E-Ent $\uparrow$ \\
    \midrule
    Ungrounded MoE & 81.42 & -- & -- & 1.38 \\
    DA-MoE w/o Process Prior & 81.76 & 0.41 & 0.88 & 1.52 \\
    DA-MoE + Process Prior & 82.28 & 0.68 & 0.43 & 1.61 \\
    \textbf{PLCD} & \textbf{83.51} & \textbf{0.72} & \textbf{0.38} & \textbf{1.67} \\
    \bottomrule
  \end{tabular}
  }
  \caption{Language-to-Cognition mapping analysis of DA-MoE expert routing.}
  \label{tab:cognitive_grounding}
  \vspace{-1.0em}
\end{table}

Table~\ref{tab:cognitive_grounding} evaluates the Language-to-Cognition Mapping module by testing whether the DA-MoE gate is aligned with the LLM-derived process prior. Detailed comparison settings and metric definitions are provided in Appendix~\ref{app:cognitive_grounding_setup}. The ungrounded MoE improves representation flexibility but provides no direct gate-prior alignment. Adding process-prior conditioning increases gate-prior cosine similarity and reduces KL divergence, indicating that the activated experts better match the cognitive operations required by the target exercise. Expert entropy also increases, suggesting that the process prior encourages broader expert utilization rather than expert collapse. Adding process-level contrastive learning further improves both prediction and grounding metrics, showing that the contrastive objective strengthens the alignment between learner evidence and process-specific exercise demands.

%% file: latex/3_experiment/Visualization.tex
\subsection{Cognitive Scaffolding in Latent Space}

\begin{figure}[t]
  \centering
  \includegraphics[width=\linewidth]{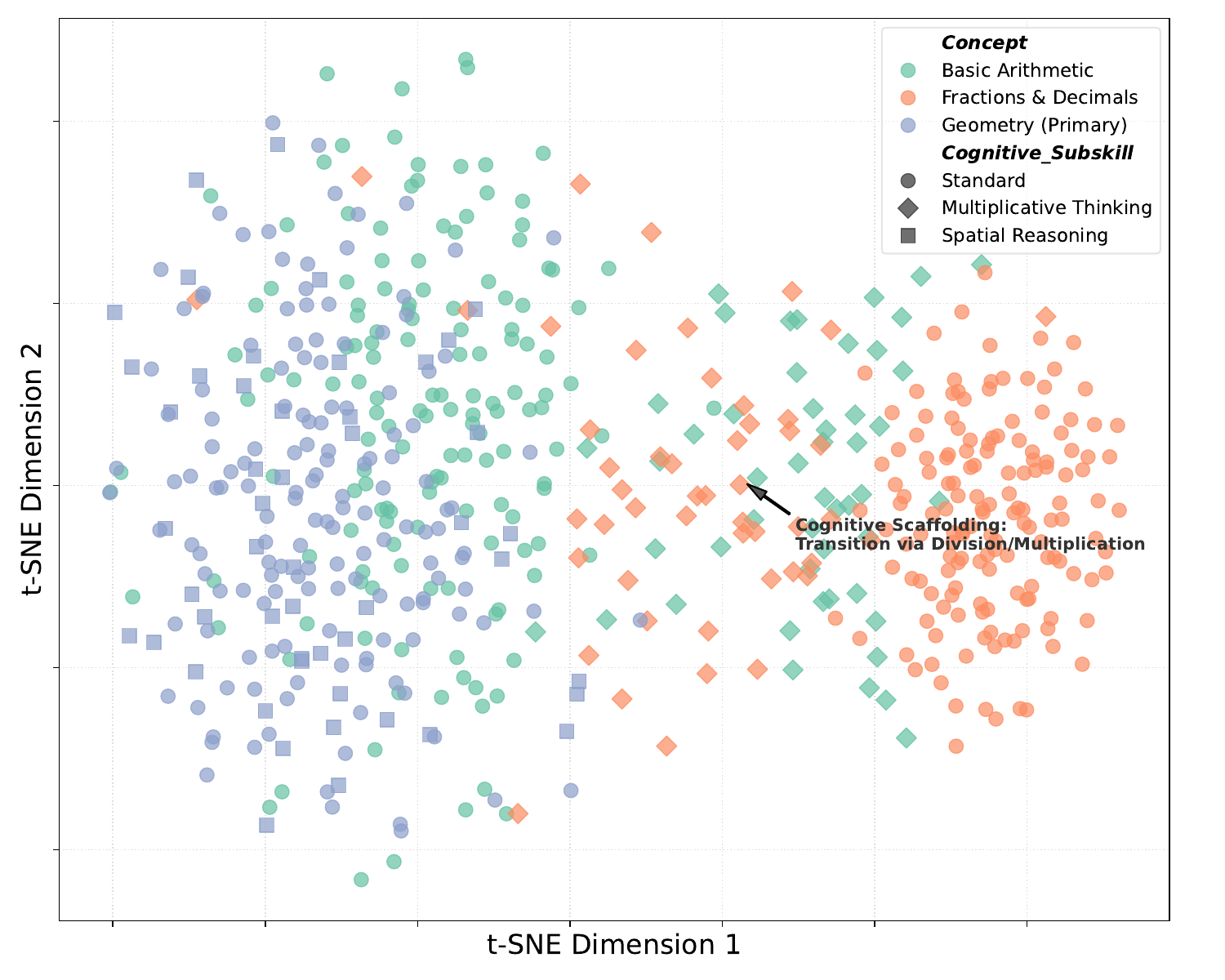}
  \caption{Latent knowledge manifold of primary mathematics concepts derived from the Junyi dataset. Each point represents a mapped exercise representation projected into two dimensions using t-SNE.}
  \vspace{-1.0em}
  \label{fig:tSNE}
\end{figure}

As an exploratory analysis, we visualize whether the learned representations exhibit structure consistent with curricular and process-level organization. We apply t-SNE~\citep{maaten2008visualizing} to project the high-dimensional exercise representations after Language-to-Cognition Mapping into a two-dimensional space. As shown in Figure~\ref{fig:tSNE}, rather than forming isolated islands, the representations form a continuous manifold. This aligns with the pedagogical reality of primary mathematics, where concepts, prerequisite relations, and problem-solving operations are highly scaffolded. We observe three phenomena that are consistent with a cognitively grounded interpretation:

\textbf{Domain-specific modularity (topic level):} Exercises belonging to distinct mathematical domains form clear clusters. For instance, geometry problems and fraction problems are localized in separate regions. This indicates that the mapped representations preserve coarse-grained curricular boundaries rather than collapsing all textual evidence into generic semantic similarity.

\textbf{Continuous skill transition (the cognitive scaffold effect):} The spatial layout reflects curriculum progression. For example, exercises related to basic arithmetic smoothly transition into fractions. Boundary regions are densely populated by exercises involving division and multiplicative thinking, which act as cognitive stepping stones.

\textbf{Latent proximity of shared processes: }Exercises from different domains cluster when they involve similar cognitive operations, such as relational mapping, spatial reasoning, or procedural manipulation. This aligns with the process-grounded DA-MoE design, where expert activation depends on reusable cognitive operations rather than dataset-specific item IDs.

%% file: latex/3_experiment/5.tex
\subsection{Diagnostic Report Analysis}

\begin{figure}[t]
    \centering
    \begin{subfigure}{0.49\linewidth}
        \centering
        \includegraphics[width=\linewidth]{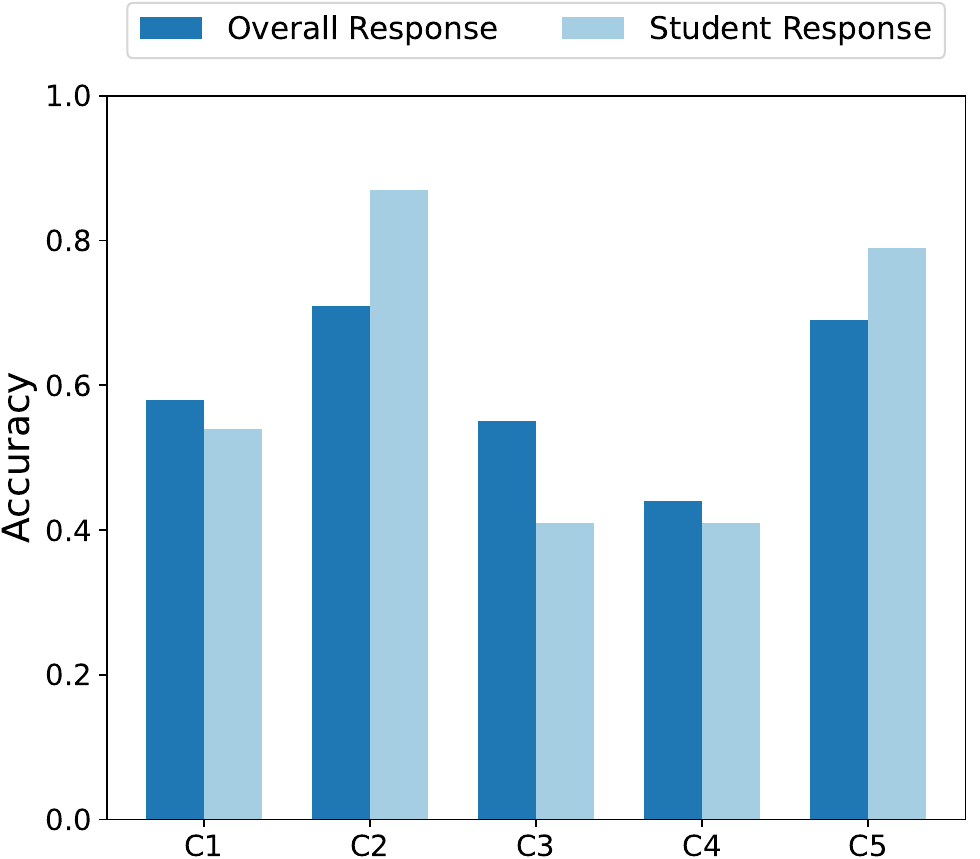}
        \caption{Correct rates.}
        \label{fig:radar_a}
    \end{subfigure}
    \hfill
    \begin{subfigure}{0.49\linewidth}
        \centering
        \includegraphics[width=\linewidth]{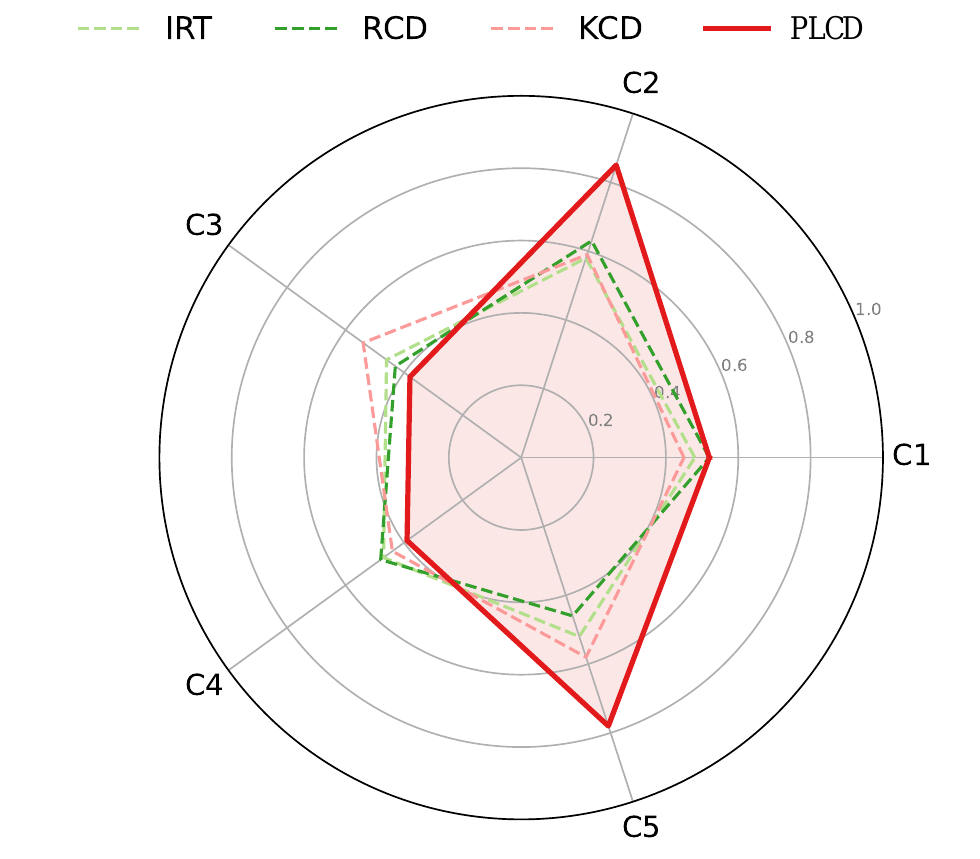}
        \caption{Diagnostic results.}
        \label{fig:radar_b}
    \end{subfigure}
    \caption{An example of a student's mastery levels diagnostic of specific concepts.}
    \vspace{-1.0em}
    \label{fig:radar}
\end{figure}

To further explore the diagnostic capability of our framework, we conduct a case study on a new student excluded from the training dataset. Five mathematical concepts from the dataset are selected for this analysis. As shown in Figure~\ref{fig:radar_a}, we first analyze the baseline correctness rate of these concepts within the general student population and compare it to the observed performance of this specific student. The behavioral data shows the student excels in $C_2$ and $C_5$, but demonstrates significant weaknesses in $C_1$, $C_3$ and $C_4$.

Next, we generate diagnostic reports for the student using different state-of-the-art baselines. As shown in Figure~\ref{fig:radar_b}, the reports generated by traditional ID-based models are relatively smooth, with predictive values confined mainly to the generic range of 0.4 to 0.6. This averaging effect indicates that models relying on discrete ID embeddings struggle to infer a fine-grained cognitive state when only limited student history is available.
In contrast, PLCD produces sharper and more distinctive diagnostic estimates. The difference comes from two sources: the concept and exercise process graphs provide structured cognitive priors, and target-conditioned memory retrieves prior responses whose cognitive operations are relevant to the target concept. This allows the diagnosis to reflect both what the exercise requires and what the student has previously demonstrated on similar process demands.%

Additional analyses are provided in the appendix: Appendix~\ref{app:calibration_analysis} studies the calibration effect of the psychometric guess--slip head. 
Appendix~\ref{app:teacher_validation} reports an independent human teacher validation of the LLM-generated cognitive process graphs.

%% file: latex/limitations.tex
\section*{Limitations}

Although PLCD shows promising results, this work is still limited in scope. The model relies on available response records to calibrate learner-specific states, and the quality of its structured cognitive evidence may be affected by the clarity of exercise text and concept descriptions. In addition, the current study evaluates cognitive diagnosis mainly through response prediction and related calibration metrics, because true learner mastery states are not directly observable in real-world datasets. The results therefore support PLCD as a predictive and cognitively grounded modeling framework, but they should not be interpreted as direct evidence of causal learning mechanisms or as a complete substitute for longitudinal educational assessment.

\section*{Ethical considerations}

This work uses publicly available and anonymized educational datasets, and does not involve collecting private student information or identifying individual learners. The proposed model is intended to support cognitive diagnosis and personalized learning analysis, rather than to replace teachers or make high-stakes educational decisions automatically. In practical use, its diagnostic results should be interpreted as model-based evidence conditioned on available response records and exercise content, and should be used with appropriate human oversight.

The teacher-validation study involved paid teacher participants. Details on participant instructions, risks, recruitment, and compensation are provided in Appendix \ref{app:teacher_validation}.

\section*{Acknowledgments}
Thanks to all reviewers, their reviews are important for this research. This work is supported by Henan provincial key research and development program (No. 241111211900) and by the Strategic Priority Research Program of the Chinese Academy of Sciences under Grant No.\ XDB0680302.

%% file: latex/appendix/sum.tex
\input{latex/appendix/related_work}
\input{latex/appendix/process_experts}
\input{latex/appendix/experimental_basic_setup}

\input{latex/appendix/cold_start}
\input{latex/appendix/memory}

\input{latex/appendix/cognitive_grounding}
\input{latex/appendix/calibration_analysis}

\input{latex/appendix/teacher_validation}

\input{latex/appendix/llm_prompts}

%% file: latex/appendix/related_work.tex
\section{Related Work}
\label{app:related_work}

\subsection{ID-based Cognitive Diagnosis Models}

Cognitive diagnosis (CD) aims to infer fine-grained learner mastery over knowledge concepts from observed responses. Classical psychometric models provide the foundation for this task. Item response theory and multidimensional item response theory model response probability through latent learner ability and item parameters \citep{embretson2013item,ackerman2003using}. Diagnostic classification models further represent mastery over discrete attributes, including DINA and its slip-guess formulation \citep{junker2001cognitive,delatorre2009dina}, DINO \citep{templin2006measurement}, G-DINA \citep{delatorre2011gdina}, LCDM \citep{henson2009lcdm}, and GDM \citep{vondavier2005gdm}. These models make diagnostic assumptions explicit and support interpretable mastery profiles, but they usually depend on manually specified Q-matrices and estimate learner/item parameters primarily from response records. As a result, their interpretability is often tied to the quality of expert-defined concept annotations, while their predictive strength depends on whether enough observations are available for reliable parameter estimation.
Recent neural CDMs relax hand-designed response functions and improve predictive accuracy by learning richer student, exercise, and concept representations. Fuzzy cognitive diagnosis models and NeuralCD-style frameworks use neural networks or fuzzy relations to model complex student-exercise interactions \citep{liu2018fuzzy,wang2020neural,wang2022neuralcd}. Relation-aware methods such as RCD construct graphs among students, exercises, and knowledge concepts to capture dependency structures \citep{gao2021rcd}. Other work enhances diagnosis with knowledge-concept relations, context features, monotonic ranking constraints, or affective states, as in KSCD, ECD, IRR, and ACD \citep{ma2022kscd,zhou2021context,tong2021irr,wang2024boosting}. Inductive and explainable learner modeling further studies identifiability and generalization under personalized diagnosis settings \citep{li2024towards}. These developments reflect a broader trend in CD research: moving from rigid psychometric response functions toward flexible representation learning while attempting to preserve diagnostic interpretability \citep{wang2024survey}.

Although these methods have substantially advanced CD, most still rely on IDs or latent vectors learned from response logs. Students, exercises, and concepts are typically represented as discrete indices whose semantics must be inferred indirectly from interaction patterns. While effective with dense historical records, such designs generalize poorly to new students, new exercises, and sparse responses. They also struggle to relate exercises unless their connections have been observed through shared concepts, shared students, or graph edges. More importantly, ID-centric embeddings provide limited access to the linguistic and conceptual structure of educational content, such as prerequisite relations, problem statements, solution operations, and common misconceptions. This weakness is especially salient in cold-start diagnosis: a new exercise may contain sufficient textual evidence for a human expert to infer the required operations, whereas an ID-based model cannot exploit it before interactions accumulate. PLCD addresses this gap by introducing language-derived structures as cognitive priors while retaining response records for personalized calibration.

\subsection{Language- and Knowledge-enhanced Cognitive Diagnosis}

A growing line of work recognizes that educational text and external knowledge can supply signals that are missing from response-only CDMs. Early text-aware diagnosis work already showed that item content can help estimate latent item properties beyond interaction counts. DIRT enhances item response theory with question text embeddings to estimate item parameters more robustly, especially for rarely observed questions \citep{cheng2019dirt}. CMNCD extends neural cognitive diagnosis to cross-modal exercises by using textual and visual item content together with prerequisite relations among concepts \citep{song2023cmncd}. TechCD improves cold-start diagnosis by transferring knowledge concept graph embeddings \citep{gao2023leveraging}. KCD uses LLM-generated knowledge to enrich cognitive diagnosis representations \citep{Dong_Chen_Wu_2025}, while DMC-CDM introduces multi-perspective consolidation through conditional diffusion \citep{zhao2025multi}. More recent LLM-based CD studies further explore open-world knowledge and cross-domain transfer. LLM4CD constructs cognitively expressive textual representations from large language models for open-world knowledge augmentation \citep{zhang2025llm4cd}, and LCST formulates zero-shot cross-domain cognitive diagnosis as an LLM-guided cognitive state transfer problem \citep{ma2025lcst}. Together, these studies indicate that semantic information can improve diagnosis when the response matrix alone is sparse or domain-bound.

Related evidence also appears in broader educational measurement. BERT-IRT shows that pretrained language representations can reduce the need for lengthy item piloting by connecting item text to interpretable IRT parameters \citep{kochmar2024bertirt}. LLM-based item difficulty and response-time estimation pipelines demonstrate that structured linguistic and semantic features can support scalable assessment development \citep{veeramani2024llmitem}. Surveys and position papers on LLMs in education further suggest that language models can help generate, interpret, and personalize educational content, while also requiring careful attention to reliability and validity \citep{kasneci2023chatgpt}. These works are not CDMs in the strict sense, but they provide useful evidence that educational language contains measurable signals about difficulty, required reasoning, and learner-facing instructional structure.

These studies show that language and external knowledge are valuable for educational diagnosis and assessment. However, many CD methods use text as an auxiliary feature for ID-based models, as an additional item-parameter estimator, or as a bridge for domain transfer. In such designs, response-log embeddings remain the main representational anchor, and language primarily enriches or regularizes those embeddings. PLCD takes a different view. It treats LLM-derived concept schemas and cognitive process graphs as the organizing cognitive evidence for diagnosis. Target-conditioned semantic memory then retrieves response records that are cognitively relevant to the current exercise, rather than compressing a learner's entire history into a fixed ID-bound profile. Finally, the process-grounded mapper aligns this language-derived evidence with diagnostic states, making the model both semantically grounded and calibrated by observed learner behavior.

%% file: latex/appendix/process_experts.tex
\section{Process Expert Inventory}
\label{app:process_expert_inventory}

Table~\ref{tab:process_experts} lists the cognitive process inventory used by the process-grounded DA-MoE mapper. The same inventory is used for Junyi, XES3G5M, and MOOC. This design treats experts as reusable cognitive operations rather than dataset-dependent categories. For each dataset, the activated expert mixture still differs across exercises through the process prior $\boldsymbol{\pi}_e$ computed from the exercise-specific cognitive process graph.

\begin{table}[h]
\centering
\small
\begin{tabular}{@{}p{0.22\linewidth}p{0.66\linewidth}@{}}
\toprule
Expert & Cognitive process \\
\midrule
$E_1$ & Semantic-to-symbolic translation \\
$E_2$ & Relational mapping \\
$E_3$ & Arithmetic computation \\
$E_4$ & Algebraic manipulation \\
$E_5$ & Spatial reasoning \\
$E_6$ & Proportional reasoning \\
$E_7$ & Conceptual recall \\
$E_8$ & Procedural execution \\
\bottomrule
\end{tabular}
\caption{Cognitive grounding of DA-MoE experts. Each expert is associated with a process type rather than an uninterpreted latent domain.}
\label{tab:process_experts}
\end{table}

%% file: latex/appendix/experimental_basic_setup.tex
\section{Experimental Basic Setup}
\label{app:experimental_basic_setup}

\subsection{Datasets}
In our experiments, we evaluate PLCD on three real-world datasets in intelligent education research, i.e., Junyi~\citep{chang2015modeling}, XES3G5M~\citep{liu2023xes3g5m}, and MOOC~\citep{yu2023moocradar}. These datasets cover different learning platforms and exhibit different scales of students, exercises, concepts, and response histories. Each dataset contains student response records together with exercise-concept correlations, which makes it possible to evaluate both response prediction and the calibration of language-derived cognitive priors by observed learner behavior. The detailed statistics are provided in Table~\ref{tab:data}.

The statistics reported in Table~\ref{tab:data} correspond to the processed experimental subsets used in this paper rather than the full raw dataset releases. Starting from the original public datasets, we filter and preprocess the response logs to retain valid student--exercise interactions with usable exercise texts and exercise--concept annotations, remove invalid or duplicate records. 

We focus on mathematics-oriented datasets deliberately. First, mathematics is one of the most representative domains for cognitive diagnosis because its knowledge structure is naturally hierarchical and prerequisite-dependent, making it suitable for evaluating whether a model can capture concept relations, procedural operations, and process-level misconceptions. Second, mathematical exercises usually have relatively objective correctness labels and explicit exercise--concept mappings, which reduces ambiguity in response interpretation and enables fair comparison with existing CDMs. Third, controlling the subject domain helps isolate the effect of our process-aware language cognitive modeling framework from confounding factors caused by cross-domain differences in discourse style, answer format, or grading criteria. Although all three datasets belong to mathematics, they are collected from different educational platforms and differ substantially in response distributions and concept structures, providing a diverse testbed for evaluating the robustness and generalizability of PLCD.

\begin{table}[t] 
\centering
\small
\resizebox{1.0\linewidth}{!}{
\begin{tabular}{l r r r}
\toprule
\textbf{Statistics} & \textbf{Junyi} & \textbf{XES3G5M} & \textbf{MOOC}\\ 
\midrule
\#Students & 10,000 & 3,000 & 3,000\\ 
\#Exercises & 835 & 4,314 & 2,726\\ 
\#Knowledge concepts & 835 & 677 & 983\\ 
\#Response records & 324,631 & 792,720 & 437,563\\ 
\#Records per student & 32.46 & 264.24 & 145.85\\ 
\bottomrule
\end{tabular}
}
\vspace{-0.3em}
\caption{The statistics of the dataset.}
\label{tab:data}
\vspace{-1.0em}
\end{table}

\vspace{-0.5em}
\subsection{Baselines and Evaluation Metrics}
\label{sec:baseline}

To verify the effectiveness of PLCD, we compare it with several representative baselines while taking their reproducibility into account. The details are presented as follows.
(1) \textbf{Traditional Cognitive Diagnosis Models}: IRT \citep{embretson2013item} predicts the correct response probabilities based on item difficulty and student ability. As an extension of IRT, MIRT \citep{ackerman2003using} further considers multiple cognitive dimensions.
(2) \textbf{Neural Cognitive Diagnosis Models}: NCD \citep{wang2020neural} directly learns student knowledge states from response data. RCD \citep{gao2021rcd} adopts relation maps to model skill dependencies and learning paths. ACD \citep{wang2024boosting} integrates affective states to improve diagnostic performance.
(3) \textbf{Knowledge-enhanced Cognitive Diagnosis Models}: KaNCD \citep{wang2022neuralcd} and TechCD \citep{gao2023leveraging} improve the generalization ability of diagnosis by incorporating external knowledge. KCD \citep{Dong_Chen_Wu_2025} leverages LLMs to enrich CDM representations, and DMC-CDM \citep{zhao2025multi} employs conditional diffusion models to boost diagnostic accuracy.

Considering that students' true mastery levels are inherently unobservable in real-world educational settings, following prior work \citep{liu2018fuzzy, ijcai2021p703}, we indirectly evaluate the effectiveness of our model by predicting students' exercise performance.
The ACC (Accuracy), AUC (Area under the curve), and RMSE (Root mean square error)~\citep{gao2021rcd} are used to assess the performance of CDMs. Because the guess-slip head produces calibrated response probabilities rather than only ranking scores, we also track Brier Score and Expected Calibration Error (ECE) in component-level calibration analysis.

\subsection{Implementation Details}

All the experiments are run on a server with dual Intel(R) Xeon Silver 4210 CPUs @ 2.20 GHz, 256 GB RAM and four NVIDIA A100 GPUs. Following the standard cognitive diagnosis setting, we split response records in each dataset into training, validation, and test sets with an 8:1:1 ratio, and use the same partitions for all models. This is a fixed cognitive diagnosis protocol rather than a knowledge-tracing protocol~\citep{liu2019ekt}: the model is trained once and is not updated sequentially during validation or testing. We retain the response order within each student only to prevent leakage in target-conditioned memory. For a validation or test target at response index $t$, PLCD can retrieve only the student's observed training responses with indices $i<t$; validation/test labels and later responses are never used as memory evidence. Hyperparameters are selected on the validation set, and the test set is used only for final reporting. All the reported results are the average of ten different runs, and statistical significance is computed across these runs.

For structured cognitive evidence construction, we employ Gemini-3 Pro as the underlying LLM~\citep{google2025gemini3pro} to generate concept schemas and cognitive process graphs. We use text-embedding-3-large~\citep{openai2024embedding3} as the text embedder for serialized schemas and process graphs. For target-conditioned semantic memory, only response records that occur before the target interaction are eligible for retrieval, preventing information leakage from future attempts. The retrieval score combines graph-level semantic similarity with Q-matrix concept overlap, controlled by $\tau_m$ and $\gamma$. For both our model and baselines, we ensure consistency by uniformly tuning shared hyperparameters on the validation set, including the embedding dimension in \{64,128,256\} and the number of negative samples in \{8,16,32\}. In the DA-MoE module, we set the number of process experts $N=8$ and the TopK gating parameter $k=2$. The process-level contrastive temperature $\tau$ is fixed at 0.1.

We train neural models using AdamW. The learning rate is selected from \{$10^{-4}$, $3\times10^{-4}$, $5\times10^{-4}$, $10^{-3}$\}, batch size from \{256,512,1024\}, and weight decay from \{0,$10^{-5}$,$10^{-4}$\}. For the memory module, $\tau_m$ is searched in \{0.05,0.1,0.2,0.5\} and $\gamma$ in \{0,0.1,0.5,1.0\}. For the overall objective, $\lambda_{1}$ for process-level contrastive learning is searched in \{0.01,0.05,0.1,0.5,1.0\}, $\lambda_{2}$ for process-prior regularization in \{0.001,0.005,0.01,0.05,0.1\}, and $\beta$ for MoE load balancing in \{0.001,0.005,0.01,0.05\}. The configuration with the best validation AUC is used for final test evaluation.

\subsection{Statistical Uncertainty}
\label{app:statistical_uncertainty}

\begin{table*}[t]
\centering
\small
\setlength{\tabcolsep}{5pt}
\renewcommand{\arraystretch}{1.1} %
\begin{tabular}{llccc}
\toprule
Dataset & Metric & Strongest baseline & Baseline & PLCD (Ours) \\
\midrule
\multirow{3}{*}{Junyi}
& ACC  & KCD     & 78.03$\pm$0.24 & 80.81$\pm$0.18 \\
& AUC  & KCD     & 82.29$\pm$0.20 & 84.52$\pm$0.16 \\
& RMSE & KCD   & 40.21$\pm$0.13 & 39.01$\pm$0.11 \\
\midrule
\multirow{3}{*}{XES3G5M}
& ACC  & KCD     & 80.01$\pm$0.21 & 83.51$\pm$0.15 \\
& AUC  & KCD     & 82.97$\pm$0.18 & 85.37$\pm$0.14 \\
& RMSE & KCD   & 39.04$\pm$0.10 & 38.13$\pm$0.09 \\
\midrule
\multirow{3}{*}{MOOC}
& ACC  & DMC-CDM & 83.81$\pm$0.19 & 87.16$\pm$0.13 \\
& AUC  & DMC-CDM & 84.31$\pm$0.17 & 88.33$\pm$0.12 \\
& RMSE & DMC-CDM & 38.91$\pm$0.10 & 36.51$\pm$0.08 \\
\bottomrule
\end{tabular}
\caption{Mean $\pm$ sample standard deviation over multiple runs for PLCD and the strongest non-PLCD baseline on student performance prediction. All values are reported in percentage points.}
\label{tab:main_std_compact}
\end{table*}

All main prediction results are averaged over ten independent runs. To quantify run-level variability without overcrowding the main table, we report the mean and sample standard deviation for PLCD and the strongest non-PLCD baseline in Table~\ref{tab:main_std_compact}. The significance marks in Table~\ref{tab:main} are computed by a two-sided paired $t$-test between PLCD and the corresponding strongest non-PLCD baseline across the same ten runs, with $p<0.05$.

%% file: latex/appendix/cold_start.tex
\section{Cold-start Evaluation Settings}
\label{app:cold_start_protocol}

We evaluate three settings that remove different sources of diagnostic evidence. 
The new-concept and new-exercise settings test whether a model can generalize to unseen entities whose response histories are unavailable during training. 
The missing-Q-matrix setting instead keeps the response records unchanged but removes the expert exercise--concept annotation channel. 
All splits and masking operations are constructed before model training, and held-out response records or masked concept links are never used for training, retrieval, or inference.
\textbf{\textit{Oracle}} denotes an upper bound trained with the full dataset and complete annotations.

\paragraph{New concepts.}
Let $\mathcal{C}_{\mathrm{new}} \subset \mathcal{C}$ denote the held-out concept set. 
We remove from training all response records $(s,e,t,y)$ whose exercise is associated with at least one held-out concept, i.e., 
$\{c \mid Q_{e,c}=1\} \cap \mathcal{C}_{\mathrm{new}} \neq \emptyset$. 
These removed records are used only for validation and testing. 
Therefore, the model cannot use training interactions to learn concept-side parameters, response patterns, or calibration signals for the held-out concepts. 
For PLCD, the textual schema of a held-out concept can still be constructed from its concept name or description, but no learner response involving the held-out concept is used to calibrate the target-conditioned memory during training. 
This setting evaluates whether language-derived concept and process evidence can support transfer to conceptual targets that lack interaction history.

\paragraph{New exercises.}
Let $\mathcal{E}_{\mathrm{new}} \subset \mathcal{E}$ denote the held-out exercise set. 
We remove all response records associated with exercises in $\mathcal{E}_{\mathrm{new}}$ from the training set and evaluate the models on these held-out exercise records. 
Thus, each target exercise in this setting has no training interaction history and no learned item-side embedding or response pattern. 
Unlike the missing-Q-matrix setting, the Q-matrix remains available in the new-exercise setting, so the evaluation focuses on exercise-level cold start rather than missing expert concept annotations. 
PLCD encodes a new exercise from its exercise text, difficulty level, available concept annotations, and LLM-derived cognitive process graph, while ID-based baselines cannot learn a reliable item representation from response logs for the held-out exercises.

\paragraph{Missing Q-matrix.}
The missing-Q-matrix setting evaluates robustness when expert exercise--concept annotations are unavailable. 
Different from the new-concept and new-exercise settings, this setting does not remove students, exercises, or response records. 
Instead, before training and inference, we mask all positive exercise--concept links in the input Q-matrix:
\begin{equation}
Q^{\mathrm{miss}}_{e,c}=0,\quad \forall (e,c)\ \mathrm{with}\ Q_{e,c}=1.
\end{equation}
All originally zero entries remain zero, so the model receives an all-zero exercise--concept matrix as input. 
The original Q-matrix is used only to construct the masked matrix and is never exposed to any model as an input feature.

For PLCD, this protocol removes every explicit Q-derived channel. 
First, the exercise graph prompt is changed from
\begin{equation}
I_e=[d_e,\{c \mid Q_{e,c}=1\},l_e]
\end{equation}
to
\begin{equation}
I^{\mathrm{miss}}_e=[d_e,\emptyset,l_e],
\end{equation}
so the LLM constructs the cognitive process graph without gold exercise--concept links. 
Second, the concept-incidence vectors used in target-conditioned retrieval are set to zero, i.e., $q_e=\mathbf{0}$ and $q_{e_i}=\mathbf{0}$, which disables the explicit concept-overlap term $q_e^\top q_{e_i}$. 
Retrieval therefore relies on graph-level semantic similarity and response-record calibration rather than Q-matrix overlap. 
Third, whenever the prediction head requires a concept set for a target exercise, PLCD uses the concept field inferred from the text-only process graph, denoted as $\widetilde{\mathcal{C}}_e$, rather than the gold concept set $\{c \mid Q_{e,c}=1\}$. 
This avoids leakage from expert annotations while allowing PLCD to exploit language-inferred cognitive evidence. 

For baselines that require a Q-matrix, we directly provide $Q^{\mathrm{miss}}$ as their input Q-matrix. Since RCD explicitly relies on relation maps constructed from student--exercise--concept associations, it cannot be properly instantiated when all positive exercise--concept links are masked. Therefore, we exclude RCD from the missing-Q-matrix setting.

%% file: latex/appendix/memory.tex
\section{Memory Mechanism Verification Setup}
\label{app:memory_setup}

To isolate the effect of the memory mechanism, we replace only the retrieval weights used to construct the learner memory vector, while keeping the remaining modules unchanged.

\subsection{Memory Strategy Variants}

\paragraph{Static Learner Summary.}
This variant replaces target-conditioned retrieval with a target-independent aggregation over the student's historical responses:
\begin{equation} \small
m^{\mathrm{static}}_{s,t}
=
\frac{1}{t-1}
\sum_{i<t}
\psi(G_{e_i}, y_i, \Delta t_i).
\end{equation}
It tests whether a general learner representation is sufficient without conditioning on the target exercise.

\paragraph{Random Retrieval.}
This variant randomly samples historical interactions from the same student history and uses the same memory budget as PLCD. 
It tests whether the improvement comes merely from adding historical responses rather than retrieving cognitively relevant evidence.

\paragraph{Concept-overlap Only.}
This variant removes graph-level semantic similarity and computes retrieval weights only from Q-matrix overlap:
\begin{equation} 
\alpha_i
=
\frac{
\exp(\gamma q_e^\top q_{e_i})
}{
\sum_{j<t}
\exp(\gamma q_e^\top q_{e_j})
}.
\end{equation}

\paragraph{Semantic Only.}
This variant removes explicit Q-matrix overlap and retrieves histories using only graph-level semantic similarity:
\begin{equation} 
\alpha_i
=
\frac{
\exp(h_e^\top h_{e_i} / \tau_m)
}{
\sum_{j<t}
\exp(h_e^\top h_{e_j} / \tau_m)
}.
\end{equation}

\paragraph{Semantic+Q Retrieval.}
This is the full PLCD memory strategy:
\begin{equation} 
\alpha_i
=
\frac{
\exp(h_e^\top h_{e_i}/\tau_m + \gamma q_e^\top q_{e_i})
}{
\sum_{j<t}
\exp(h_e^\top h_{e_j}/\tau_m + \gamma q_e^\top q_{e_j})
}.
\end{equation}

\paragraph{Semantic+Q w/ Shuffled Labels.}
This variant uses the same retrieval weights and retrieved historical exercises as Semantic+Q retrieval. 
However, the correctness label $y_i$ in each historical memory item is replaced by a shuffled label $\tilde{y}_i$, obtained by randomly permuting correctness labels within the same student's historical prefix. 
This preserves the student's marginal correctness distribution but breaks the alignment between each retrieved exercise and the student's actual response to that exercise:
\begin{equation} 
m^{\mathrm{shuffle}}_{s,e,t}
=
\sum_{i<t}
\alpha_i
\cdot
\psi(G_{e_i}, \tilde{y}_i, \Delta t_i).
\end{equation}
Random retrieval and label shuffling are averaged over five random seeds.

\subsection{Retrieval-level Diagnostics.}
\label{app:retrieval_diagnostics}

In addition to prediction metrics, we report two retrieval-level diagnostics to examine whether the memory module retrieves cognitively and behaviorally meaningful histories.

\paragraph{Q-Jaccard@K.}
Q-Jaccard@K measures the average concept overlap between the target exercise and the top-$K$ retrieved historical exercises according to the Q-matrix:

\begin{small}
\begin{multline*}
\mathrm{Q\text{-}Jac.@}K
=
\frac{1}{|\mathcal{T}|}
\sum_{(s,e,t)\in\mathcal{T}}
\frac{1}{K}
\sum_{i\in R_K(s,e,t)} \\
\times
\frac{
q_e^\top q_{e_i}
}{
\|q_e\|_1 + \|q_{e_i}\|_1 - q_e^\top q_{e_i}
}.
\end{multline*}
\end{small}
Here, $\mathcal{T}$ denotes the evaluation set, and $R_K(s,e,t)$ denotes the top-$K$ historical interactions retrieved for target interaction $(s,e,t)$. 
A higher Q-Jaccard@K indicates that the memory module retrieves histories that are more conceptually aligned with the target exercise.

\paragraph{Residual Response Correlation@K.}
Residual Response Correlation@K measures whether the retrieved historical responses are predictive of the target response after removing the student's overall prior correctness level. 
For each target interaction $(s,e,t)$, we first compute the student's historical average correctness before time $t$:
\begin{equation} \small
\bar{y}_{s,<t}
=
\frac{1}{t-1}
\sum_{j<t} y_{s,e_j,t_j}.
\end{equation}
We then compute the weighted residual correctness of the retrieved histories:
\begin{equation} 
r^{\mathrm{mem}}_{s,e,t}
=
\sum_{i\in R_K(s,e,t)}
\alpha_i
\left(
y_{s,e_i,t_i} - \bar{y}_{s,<t}
\right),
\end{equation}
and the residual target correctness:
\begin{equation} 
r^{\mathrm{tar}}_{s,e,t}
=
y_{s,e,t} - \bar{y}_{s,<t}.
\end{equation}
Residual Response Correlation@K, computed as the Pearson correlation coefficient:
\begin{small}
\begin{equation}
\mathrm{Res.\ Corr.@}K
=
\mathrm{P}
\left(
\{r^{\mathrm{tar}}_{s,e,t}\}_{(s,e,t)\in\mathcal{T}},
\{r^{\mathrm{mem}}_{s,e,t}\}_{(s,e,t)\in\mathcal{T}}
\right).
\end{equation}
\end{small}

This metric tests whether the retrieved historical correctness labels provide learner-specific behavioral evidence beyond the student's marginal ability level. 
For the label-shuffled memory variant, the retrieved exercises remain unchanged, so Q-Jaccard@K is preserved, but Residual Response Correlation@K should drop because the alignment between exercises and correctness labels is corrupted.

%% file: latex/appendix/cognitive_grounding.tex
\section{Cognitive Grounding Evaluation Setup}
\label{app:cognitive_grounding_setup}

This appendix details the evaluation protocol for the cognitive grounding analysis in Table~\ref{tab:cognitive_grounding}. Across all variants, we only vary whether DA-MoE routing uses the LLM-derived process prior and whether process-level supervised contrastive learning is applied.

\subsection{Compared Variants}

\paragraph{Ungrounded MoE.}
This variant uses a standard mixture-of-experts layer. The experts are not bound to predefined cognitive process types, and the routing gate does not use the LLM-derived process prior. It tests whether a flexible MoE alone is sufficient to explain the gains.

\paragraph{DA-MoE w/o Process Prior.}
This variant keeps the expert inventory aligned with cognitive process types, but removes the process-prior signal from the routing gate. Concretely, the gate is computed only from the input representation and does not inject $\log(\boldsymbol{\pi}_e)$ or use the process-prior KL anchor. It tests whether process-labeled experts are useful without explicit prior-guided routing.

\paragraph{DA-MoE + Process Prior.}
This variant injects the exercise process prior $\boldsymbol{\pi}_e$ into the DA-MoE gate and applies process-prior regularization. It evaluates whether expert routing becomes more cognitively aligned when the activated experts are encouraged to match the process demands inferred from the exercise graph.

\paragraph{PLCD (+ Process-level CL).}
This is the full PLCD variant. In addition to process-prior-guided DA-MoE routing, it uses process-level supervised contrastive learning to align learner evidence with process-specific exercise demands. This setting tests whether process-level contrastive supervision provides additional grounding beyond gate regularization.

\subsection{Evaluation Metrics}

\paragraph{Gate-Prior Cosine (GP-Cos).}
Gate-Prior Cosine measures the average cosine similarity between the DA-MoE gate distribution $\mathbf{g}_e$ and the LLM-derived process prior $\boldsymbol{\pi}_e$:
\begin{equation}
\mathrm{Cos}
=
\frac{1}{|\mathcal{E}|}
\sum_{e\in\mathcal{E}}
\frac{\mathbf{g}_e^\top \boldsymbol{\pi}_e}
{\|\mathbf{g}_e\|_2 \|\boldsymbol{\pi}_e\|_2}.
\end{equation}
A higher value indicates that the routing distribution is more consistent with the process demand inferred from the exercise.

\paragraph{KL($\boldsymbol{\pi}_e \| \mathbf{g}_e$).}
This metric measures the average KL divergence from the process prior to the gate distribution:
\begin{equation}
\mathrm{KL}
=
\frac{1}{|\mathcal{E}|}
\sum_{e\in\mathcal{E}}
\sum_{p\in\mathcal{P}}
\pi_{e,p}
\log
\frac{\pi_{e,p}+\varepsilon}{g_{e,p}+\varepsilon}.
\end{equation}
A lower value indicates that the DA-MoE gate better follows the cognitive process prior. For Ungrounded MoE, Gate-Prior Cosine and KL are reported as ``--'' because its experts do not have stable process semantics.

\paragraph{Expert Entropy (E-Ent).}
Expert Entropy measures the entropy of the gate distribution:
\begin{equation}
\small
\mathrm{Ent}
=
\frac{1}{|\mathcal{E}|}
\sum_{e\in\mathcal{E}}
\left(
-\sum_{p\in\mathcal{P}} g_{e,p}\log(g_{e,p}+\varepsilon)
\right).
\end{equation}
A higher value indicates broader expert utilization and helps diagnose whether routing collapses to a small number of experts.

%% file: latex/appendix/calibration_analysis.tex
\section{Calibration Analysis}
\label{app:calibration_analysis}

This appendix evaluates whether the psychometric guess-slip head improves the calibration of PLCD's predicted response probabilities. We compare three response heads: \textit{w/o Guess-Slip}, which directly uses the prediction probability produced by the diagnostic state; \textit{Global Guess-Slip}, which applies a dataset-level guessing and slipping correction shared by all exercises; and \textit{Exercise-level Guess-Slip (PLCD)}, which estimates exercise-specific guessing and slipping effects from the exercise representation. For a student-exercise pair, the corrected probability follows the standard psychometric form $\hat{y}=g+(1-g-s)p$, where $p$ is the uncorrected correctness probability and $g$ and $s$ denote guessing and slipping factors. We denote the three response heads as $\mathcal{H}_{0}$, $\mathcal{H}_{g}$, and $\mathcal{H}_{e}$, corresponding to \textit{w/o Guess-Slip}, \textit{Global Guess-Slip}, and \textit{Exercise-level Guess-Slip}, respectively.
We report both discrimination and calibration metrics. ACC measures thresholded prediction accuracy. Brier Score is the mean squared error between predicted probabilities and binary outcomes. ECE partitions predictions into confidence bins and measures the gap between predicted confidence and empirical accuracy. NLL is the negative log-likelihood of the observed responses under the predicted probabilities and penalizes overconfident incorrect predictions.

\begin{table}[t]
  \centering
  \small
  \resizebox{1.0\linewidth}{!}{
  \begin{tabular}{llcccc}
    \toprule
    \textbf{Dataset} & \textbf{Head} & ACC\%$\uparrow$ & Brier$\downarrow$ & ECE$\downarrow$ & NLL$\downarrow$ \\
    \midrule
    \multirow{3}{*}{XES3G5M}
    & $\mathcal{H}_{0}$ & 82.09 & 0.165 & 0.071 & 0.514 \\
    & $\mathcal{H}_{g}$ & 82.38 & 0.158 & 0.052 & 0.501 \\
    & $\mathcal{H}_{e}$ & \textbf{83.51} & \textbf{0.145} & \textbf{0.037} & \textbf{0.486} \\
    \midrule
    \multirow{3}{*}{Junyi}
    & $\mathcal{H}_{0}$ & 79.46 & 0.178 & 0.083 & 0.546 \\
    & $\mathcal{H}_{g}$ & 79.73 & 0.171 & 0.061 & 0.532 \\
    & $\mathcal{H}_{e}$ & \textbf{80.81} & \textbf{0.152} & \textbf{0.044} & \textbf{0.515} \\
    \midrule
    \multirow{3}{*}{MOOC}
    & $\mathcal{H}_{0}$ & 86.43 & 0.139 & 0.064 & 0.412 \\
    & $\mathcal{H}_{g}$ & 86.71 & 0.137 & 0.047 & 0.401 \\
    & $\mathcal{H}_{e}$ & \textbf{87.16} & \textbf{0.133} & \textbf{0.033} & \textbf{0.386} \\
    \bottomrule
  \end{tabular}
  }
  \caption{Calibration analysis of the psychometric guess-slip head. 
  $\mathcal{H}_{0}$, $\mathcal{H}_{g}$, and $\mathcal{H}_{e}$ denote \textit{w/o Guess-Slip}, \textit{Global Guess-Slip}, and \textit{Exercise-level Guess-Slip}.}
  \label{tab:calibration}
  \vspace{-1.0em}
\end{table}

Table~\ref{tab:calibration} shows that adding guess-slip parameters consistently improves performance across all three datasets. While the ACC gains are modest, the reductions in Brier Score, ECE, and NLL are more pronounced, indicating that the psychometric head mainly improves probability calibration rather than only thresholded accuracy.

Moreover, $\mathcal{H}_{e}$ consistently outperforms $\mathcal{H}_{g}$, suggesting that guessing and slipping are not uniform dataset-level effects. 
Instead, different exercises induce different opportunities for accidental success or failure due to variations in difficulty, format, distractor quality, and cognitive demand. By estimating exercise-specific guessing and slipping factors, PLCD can better model such response noise and produce more reliable probabilistic predictions.

%% file: latex/appendix/teacher_validation.tex
\section{Human Teacher Validation of LLM-Generated Cognitive Process Graphs}
\label{app:teacher_validation}

This appendix reports a teacher-validation study assessing whether LLM-generated cognitive process graphs are educationally plausible.
The numerical results in this subsection are based on teacher ratings.

\paragraph{Evaluation protocol.}
Five teachers with mathematics teaching and educational measurement backgrounds independently evaluated a stratified sample of exercises.
We sampled 40 exercises from each dataset, Junyi, XES3G5M, and MOOC, resulting in 120 exercises in total.
The sampling procedure stratifies by dataset, difficulty level, knowledge concept, and dominant process type so that the evaluated examples cover both routine and cognitively heterogeneous exercises.

For each exercise, teachers were shown the exercise text, the Q-matrix-linked concepts, the difficulty level, and the LLM-generated cognitive process graph.
They were not shown student response records, PLCD predictions, baseline outputs, or model performance.
This separation ensures that the validation focuses on the quality of the structured cognitive evidence rather than on downstream prediction accuracy.
To examine the effect of the generator model, we used the same prompt templates and sampling protocol to compare GPT-5~\citep{openai2025gpt5}, Gemini 3 Pro~\citep{google2025gemini3pro}, Qwen3-32B~\citep{yang2025qwen3}, Llama-3-70B~\citep{dubey2024llama}, and Claude Sonnet 4~\citep{anthropic2025sonnet4}.
The generated graphs were anonymized by model identity before teacher rating.

\paragraph{Human subjects and participant information.}
The appendix experiment involved real teachers as paid human participants. Before the study, participants were informed of the experiment purpose, task requirements, voluntary nature of participation, right to withdraw, and the instruction not to include personally identifiable student information. The study posed minimal risk, mainly limited to time commitment and possible fatigue from reviewing examples. Teachers were recruited through education-related professional or academic networks based on relevant teaching experience, and were compensated at a reasonable rate according to the expected workload and local context.

\paragraph{Rating dimensions.}
Teachers rated each cognitive process graph on five dimensions using a 1--5 Likert scale:
(1) concept relevance, measuring whether the graph identifies the required knowledge concepts;
(2) operation correctness, measuring whether the listed cognitive operations match the reasoning required by the exercise;
(3) solution-step completeness, measuring whether the graph covers the main solution steps;
(4) misconception plausibility, measuring whether the listed potential errors are pedagogically reasonable; and
(5) pedagogical usefulness, measuring whether the graph would help a teacher interpret the exercise's cognitive demands.
Teachers also selected the top two cognitive process labels from the fixed process inventory in Appendix~\ref{app:process_expert_inventory}.
These labels are used to compare teacher judgments with the process labels and weights produced by the LLM-generated graph.

\paragraph{Aggregation and agreement computation.}
For each rating dimension, we average scores across teachers and exercises and report the mean and standard deviation.
For process-label agreement, each teacher's selected labels are compared with the highest-weighted process labels in the LLM-generated graph.
Top-1 agreement measures exact agreement on the dominant process label, while Top-2 agreement measures whether the teacher-selected process set overlaps with the two highest-weighted LLM process labels.
Weighted process-prior cosine compares the teacher-derived process distribution with the LLM-derived process-prior vector.
Fleiss' kappa measures inter-rater agreement over coarse process labels.

\begin{table}[t]
  \centering
  \small
  \resizebox{1.0\linewidth}{!}{
  \begin{tabular}{lcccc}
    \toprule
    \textbf{Generator} & \textbf{Overall} & \textbf{Top-1} & \textbf{Top-2} & \textbf{Cosine} \\
    \midrule
    GPT-5 & 4.39 & 0.74 & 0.86 & 0.83 \\
    Gemini 3 Pro & \textbf{4.44} & \textbf{0.78} & \textbf{0.89} & \textbf{0.86} \\
    Qwen3-32B & 4.26 & 0.71 & 0.84 & 0.81 \\
    Llama-3-70B & 4.08 & 0.66 & 0.79 & 0.76 \\
    Claude Sonnet 4 & 4.31 & 0.72 & 0.85 & 0.82 \\
    \bottomrule
  \end{tabular}
  }
  \caption{Teacher-validation comparison of different LLM generators. Overall denotes the mean score across the five rating dimensions.}
  \label{tab:llm_generator_comparison}
\end{table}

\begin{table}[t]
  \centering
  \small
  \resizebox{1.0\linewidth}{!}{
  \begin{tabular}{lc}
    \toprule
    \textbf{Rating dimension} & \textbf{Score} \\
    \midrule
    Concept relevance & $4.58 \pm 0.31$ \\
    Operation correctness & $4.51 \pm 0.35$ \\
    Solution-step completeness & $4.39 \pm 0.42$ \\
    Misconception plausibility & $4.23 \pm 0.48$ \\
    Pedagogical usefulness & $4.47 \pm 0.34$ \\
    \bottomrule
  \end{tabular}
  }
  \caption{Teacher-validation scores for cognitive process graphs generated by LLM.}
  \label{tab:teacher_validation_scores}
\end{table}

\begin{table}[t]
  \centering
  \small
  \resizebox{1.0\linewidth}{!}{
  \begin{tabular}{lc}
    \toprule
    \textbf{Agreement metric} & \textbf{Value} \\
    \midrule
    Top-1 process agreement & $0.78$ \\
    Top-2 process agreement & $0.89$ \\
    Weighted process-prior cosine & $0.86$ \\
    Fleiss' kappa over coarse process labels & $0.72$ \\
    \bottomrule
  \end{tabular}
  }
  \caption{Agreement between teacher-selected process labels and LLM-generated process labels.}
  \label{tab:teacher_process_agreement}
\end{table}

\paragraph{Interpretation.}
The generator comparison in Table~\ref{tab:llm_generator_comparison} shows that Gemini 3 Pro obtains the highest overall teacher score and the strongest process-label agreement.
We therefore use Gemini 3 Pro as the default generator for constructing concept schemas and exercise cognitive process graphs.
The detailed Gemini 3 Pro results in Tables~\ref{tab:teacher_validation_scores} and~\ref{tab:teacher_process_agreement} show that teachers generally rated its generated cognitive process graphs as educationally plausible.
The high scores on concept relevance and operation correctness support the claim that the LLM-generated graphs capture teacher-recognizable cognitive demands.
The relatively lower score for misconception plausibility reflects an expected limitation: possible error types are harder to infer from exercise text alone and may require teacher correction before high-stakes deployment.
These results serve as an external plausibility check for the structured cognitive priors, complementing the model-side grounding analysis in Appendix~\ref{app:cognitive_grounding_setup}.

%% file: latex/appendix/llm_prompts.tex
\section{LLM Prompt Templates for Structured Cognitive Evidence}
\label{app:llm_prompts}
This appendix reports the prompt templates used for the LLM calls that construct cognitive priors before response-based calibration. The templates specify the reproducible instructions, but are not full API logs.

\subsection{Concept Schema Prompt}
\label{app:prompt_concept_schema}

The concept prompt $M_c$ is instantiated with the concept name $I_c=\operatorname{name}(c)$, as shown in Figure~\ref{fig:prompt_concept_schema}.

\begin{figure*}[t]
\centering
\begin{lcdpanel}[width=0.96\textwidth]{LCDBlueBack}{LCDBlueFrame}{Concept schema prompt $M_c$}
\footnotesize
\ttfamily
\textbf{System:} You are an expert educational measurement analyst. Construct structured cognitive evidence for cognitive diagnosis. Use only educationally supported facts. Avoid unsupported assumptions. If a field cannot be inferred from the given concept name, write unknown. Return deterministic plain text with the requested fields and no extra commentary.\par
\vspace{0.25em}
\textbf{User:} Concept name: [CONCEPT NAME]. Construct a concept schema with the following fields.\par
1. definition: one concise sentence describing the concept.\par
2. prerequisites: prerequisite concepts or unknown.\par
3. subskills: fine-grained skills needed to apply the concept.\par
4. common misconceptions: likely learner errors or unknown.\par
5. related cognitive operations: operations needed to use the concept, aligned with the process inventory when possible.\par
6. notes: short clarification only if needed; otherwise unknown.
\end{lcdpanel}
\vspace{0.25em}
\caption{Prompt template used to generate the concept schema $\mathcal{G}_c$. The placeholder is replaced with the target concept name.}
\label{fig:prompt_concept_schema}
\end{figure*}

\subsection{Exercise Cognitive Process Graph Prompt}
\label{app:prompt_exercise_graph}

The exercise prompt $M_e$ is instantiated with the exercise text, the Q-matrix concept set, and the difficulty level, i.e., $I_e=[d_e,\{c \mid Q_{e,c}=1\},l_e]$, as shown in Figure~\ref{fig:prompt_exercise_graph}.

\begin{figure*}[t]
\centering
\begin{lcdpanel}[width=0.96\textwidth]{LCDAmberBack}{LCDAmberFrame}{Exercise cognitive process graph prompt $M_e$}
\footnotesize
\ttfamily
\textbf{System:} You are an expert educational measurement analyst. Build a cognitive process graph for a single exercise. Ground every field in the exercise text, the provided Q-matrix concepts, and the difficulty level. Do not use learner response records. Avoid unsupported assumptions. If evidence is missing, write unknown. Return deterministic plain text with the requested fields and no extra commentary.\par
\vspace{0.25em}
\textbf{User:} Exercise text: [EXERCISE TEXT]. Q-matrix concepts: [CONCEPT LIST]. Difficulty level: [DIFFICULTY LEVEL]. Construct a cognitive process graph with the following fields.\par
1. required concepts: list each required concept with an importance weight in [0,1]; weights should sum to 1 when possible.\par
2. cognitive operations: list each operation with an importance weight in [0,1] and map it to one of the following experts: E1 semantic-to-symbolic translation; E2 relational mapping; E3 arithmetic computation; E4 algebraic manipulation; E5 spatial reasoning; E6 proportional reasoning; E7 conceptual recall; E8 procedural execution.\par
3. textual cues: quote or paraphrase exercise phrases that trigger the required concepts or operations.\par
4. solution steps: concise, observable steps needed to solve the exercise without revealing unnecessary reasoning detail.\par
5. potential error types: misconceptions, process-level failures, or careless mistakes that could plausibly lead to an incorrect answer.\par
6. constraints: preserve the exercise meaning, keep labels consistent across exercises, and use unknown for unsupported fields.
\end{lcdpanel}
\vspace{0.25em}
\caption{Prompt template used to generate the exercise cognitive process graph $\mathcal{G}_e$. The placeholders are replaced with the exercise text, Q-matrix concepts, and difficulty level.}
\label{fig:prompt_exercise_graph}
\end{figure*}

%% file: custom.bib
@inproceedings{ijcai2021p703,
  title     = {Towards a New Generation of Cognitive Diagnosis},
  author    = {Liu, Qi},
  booktitle = {Proceedings of the Thirtieth International Joint Conference on
               Artificial Intelligence, {IJCAI-21}},
  editor    = {Zhi-Hua Zhou},
  pages     = {4961--4964},
  year      = {2021},
  month     = {8},
  doi       = {10.24963/ijcai.2021/703},
  url       = {https://doi.org/10.24963/ijcai.2021/703},
}

@article{yang2025qwen3,
  title={Qwen3 technical report},
  author={Yang, An and Li, Anfeng and Yang, Baosong and Zhang, Beichen and Hui, Binyuan and Zheng, Bo and Yu, Bowen and Gao, Chang and Huang, Chengen and Lv, Chenxu},
  journal={arXiv preprint arXiv:2505.09388},
  year={2025}
}

@article{dubey2024llama,
  title={The llama 3 herd of models},
  author={Grattafiori, Aaron and Dubey, Abhimanyu and Jauhri, Abhinav and Pandey, Abhinav and Kadian, Abhishek and Al-Dahle, Ahmad and Letman, Aiesha and Mathur, Akhil and Schelten, Alan and Vaughan, Alex and others},
  journal={arXiv preprint arXiv:2407.21783},
  year={2024}
}

@misc{openai2025gpt5,
  title={{GPT-5}},
  author={{OpenAI}},
  year={2025},
  howpublished={\url{https://openai.com/gpt-5}}
}

@misc{google2025gemini3pro,
  title={{Gemini 3 Pro}},
  author={{Google DeepMind}},
  year={2025},
  howpublished={\url{https://deepmind.google/en/models/gemini/pro/}}
}

@book{anderson2013architecture,
  title={The architecture of cognition},
  author={Anderson, John R},
  year={2013},
  publisher={Psychology Press}
}

@article{chi1981categorization,
  title={Categorization and representation of physics problems by experts and novices},
  author={Chi, Michelene TH and Feltovich, Paul J and Glaser, Robert},
  journal={Cognitive science},
  volume={5},
  number={2},
  pages={121--152},
  year={1981},
  publisher={Elsevier}
}

@inproceedings{khajah2014integrating,
  title={Integrating latent-factor and knowledge-tracing models to predict individual differences in learning.},
  author={Khajah, Mohammad and Wing, Rowan and Lindsey, Robert V and Mozer, Michael},
  booktitle={Edm},
  pages={99--106},
  year={2014},
  organization={London}
}

@article{liu2023xes3g5m,
  title={Xes3g5m: A knowledge tracing benchmark dataset with auxiliary information},
  author={Liu, Zitao and Liu, Qiongqiong and Guo, Teng and Chen, Jiahao and Huang, Shuyan and Zhao, Xiangyu and Tang, Jiliang and Luo, Weiqi and Weng, Jian},
  journal={Advances in Neural Information Processing Systems},
  volume={36},
  pages={32958--32970},
  year={2023}
}

@inproceedings{chang2015modeling,
  title={Modeling exercise relationships in E-learning: A unified approach.},
  author={Chang, Haw-Shiuan and Hsu, Hwai-Jung and Chen, Kuan-Ta},
  booktitle={EDM},
  pages={532--535},
  year={2015}
}

@book{embretson2013item,
  title={Item response theory for psychologists},
  author={Embretson, Susan E and Reise, Steven P},
  year={2013},
  publisher={Psychology Press}
}

@article{ackerman2003using,
  title={Using multidimensional item response theory to evaluate educational and psychological tests},
  author={Ackerman, Terry A and Gierl, Mark J and Walker, Cindy M},
  journal={Educational Measurement: Issues and Practice},
  volume={22},
  number={3},
  pages={37--51},
  year={2003},
  publisher={Wiley Online Library}
}

@inproceedings{wang2020neural,
  title={Neural cognitive diagnosis for intelligent education systems},
  author={Wang, Fei and Liu, Qi and Chen, Enhong and Huang, Zhenya and Chen, Yuying and Yin, Yu and Huang, Zai and Wang, Shijin},
  booktitle={Proceedings of the AAAI conference on artificial intelligence},
  volume={34},
  number={04},
  pages={6153--6161},
  year={2020}
}

@inproceedings{gao2021rcd,
  title={RCD: Relation map driven cognitive diagnosis for intelligent education systems},
  author={Gao, Weibo and Liu, Qi and Huang, Zhenya and Yin, Yu and Bi, Haoyang and Wang, Mu-Chun and Ma, Jianhui and Wang, Shijin and Su, Yu},
  booktitle={Proceedings of the 44th international ACM SIGIR conference on research and development in information retrieval},
  pages={501--510},
  year={2021}
}

@inproceedings{wang2024boosting,
  title={Boosting neural cognitive diagnosis with student’s affective state modeling},
  author={Wang, Shanshan and Zeng, Zhen and Yang, Xun and Xu, Ke and Zhang, Xingyi},
  booktitle={Proceedings of the AAAI Conference on Artificial Intelligence},
  volume={38},
  number={1},
  pages={620--627},
  year={2024}
}

@article{wang2022neuralcd,
  title={NeuralCD: a general framework for cognitive diagnosis},
  author={Wang, Fei and Liu, Qi and Chen, Enhong and Huang, Zhenya and Yin, Yu and Wang, Shijin and Su, Yu},
  journal={IEEE Transactions on Knowledge and Data Engineering},
  volume={35},
  number={8},
  pages={8312--8327},
  year={2023},
  publisher={IEEE}
}

@inproceedings{gao2023leveraging,
  title={Leveraging transferable knowledge concept graph embedding for cold-start cognitive diagnosis},
  author={Gao, Weibo and Wang, Hao and Liu, Qi and Wang, Fei and Lin, Xin and Yue, Linan and Zhang, Zheng and Lv, Rui and Wang, Shijin},
  booktitle={Proceedings of the 46th international ACM SIGIR conference on research and development in information retrieval},
  pages={983--992},
  year={2023}
}

@article{Dong_Chen_Wu_2025, title={Knowledge Is Power: Harnessing Large Language Models for Enhanced Cognitive Diagnosis}, volume={39}, 
number={1}, journal={Proceedings of the AAAI Conference on Artificial Intelligence}, author={Dong, Zhiang and Chen, Jingyuan and Wu, Fei}, year={2025}, month={Apr.}, pages={164-172} }

@inproceedings{zhao2025multi,
  title={Multi-Perspective Consolidation Enhanced Cognitive Diagnosis via Conditional Diffusion Model},
  author={Zhao, Guanhao and Huang, Zhenya and Cheng, Cheng and Zhuang, Yan and Mao, Qingyang and Li, Xin and Wang, Shijin and Chen, Enhong},
  booktitle={Proceedings of the AAAI Conference on Artificial Intelligence},
  volume={39},
  number={1},
  pages={1174--1182},
  year={2025}
}

@article{liu2018fuzzy,
  title={Fuzzy cognitive diagnosis for modelling examinee performance},
  author={Liu, Qi and Wu, Runze and Chen, Enhong and Xu, Guandong and Su, Yu and Chen, Zhigang and Hu, Guoping},
  journal={ACM Transactions on Intelligent Systems and Technology (TIST)},
  volume={9},
  number={4},
  pages={1--26},
  year={2018},
  publisher={ACM New York, NY, USA}
}

@inproceedings{yu2023moocradar,
  title={Moocradar: A fine-grained and multi-aspect knowledge repository for improving cognitive student modeling in moocs},
  author={Yu, Jifan and Lu, Mengying and Zhong, Qingyang and Yao, Zijun and Tu, Shangqing and Liao, Zhengshan and Li, Xiaoya and Li, Manli and Hou, Lei and Zheng, Hai-Tao},
  booktitle={Proceedings of the 46th International ACM SIGIR Conference on Research and Development in Information Retrieval},
  pages={2924--2934},
  year={2023}
}

@article{liu2019ekt,
  title={Ekt: Exercise-aware knowledge tracing for student performance prediction},
  author={Liu, Qi and Huang, Zhenya and Yin, Yu and Chen, Enhong and Xiong, Hui and Su, Yu and Hu, Guoping},
  journal={IEEE Transactions on Knowledge and Data Engineering},
  volume={33},
  number={1},
  pages={100--115},
  year={2019},
  publisher={IEEE}
}

@inproceedings{anderson2014engaging,
  title={Engaging with massive online courses},
  author={Anderson, Ashton and Huttenlocher, Daniel and Kleinberg, Jon and Leskovec, Jure},
  booktitle={Proceedings of the 23rd international conference on World wide web},
  pages={687--698},
  year={2014}
}

@article{wang2024survey,
  title={A survey of models for cognitive diagnosis: New developments and future directions},
  author={Wang, Fei and Gao, Weibo and Liu, Qi and Li, Jiatong and Zhao, Guanhao and Zhang, Zheng and Huang, Zhenya and Zhu, Mengxiao and Wang, Shijin and Tong, Wei},
  journal={arXiv preprint arXiv:2407.05458},
  year={2024}
}

@inproceedings{li2024towards,
  title={Towards the identifiability and explainability for personalized learner modeling: an inductive paradigm},
  author={Li, Jiatong and Liu, Qi and Wang, Fei and Liu, Jiayu and Huang, Zhenya and Yao, Fangzhou and Zhu, Linbo and Su, Yu},
  booktitle={Proceedings of the ACM Web Conference 2024},
  pages={3420--3431},
  year={2024}
}

@article{maaten2008visualizing,
  title={Visualizing data using t-SNE},
  author={Maaten, Laurens van der and Hinton, Geoffrey},
  journal={Journal of machine learning research},
  volume={9},
  number={Nov},
  pages={2579--2605},
  year={2008}
}

@article{junker2001cognitive,
  title={Cognitive assessment models with few assumptions, and connections with nonparametric item response theory},
  author={Junker, Brian W and Sijtsma, Klaas},
  journal={Applied Psychological Measurement},
  volume={25},
  number={3},
  pages={258--272},
  year={2001},
  publisher={Sage Publications}
}

@article{delatorre2009dina,
  title={DINA model and parameter estimation: A didactic},
  author={de la Torre, Jimmy},
  journal={Journal of Educational and Behavioral Statistics},
  volume={34},
  number={1},
  pages={115--130},
  year={2009},
  doi={10.3102/1076998607309474},
  publisher={Sage Publications}
}

@article{templin2006measurement,
  title={Measurement of psychological disorders using cognitive diagnosis models},
  author={Templin, Jonathan L. and Henson, Robert A.},
  journal={Psychological Methods},
  volume={11},
  number={3},
  pages={287--305},
  year={2006},
  doi={10.1037/1082-989X.11.3.287},
  publisher={American Psychological Association}
}

@article{delatorre2011gdina,
  title={The generalized {DINA} model framework},
  author={de la Torre, Jimmy},
  journal={Psychometrika},
  volume={76},
  number={2},
  pages={179--199},
  year={2011},
  doi={10.1007/s11336-011-9207-7},
  publisher={Springer}
}

@article{henson2009lcdm,
  title={Defining a family of cognitive diagnosis models using log-linear models with latent variables},
  author={Henson, Robert A. and Templin, Jonathan L. and Willse, John T.},
  journal={Psychometrika},
  volume={74},
  number={2},
  pages={191--210},
  year={2009},
  doi={10.1007/s11336-008-9089-5},
  publisher={Springer}
}

@techreport{vondavier2005gdm,
  title={A general diagnostic model applied to language testing data},
  author={von Davier, Matthias},
  institution={Educational Testing Service},
  number={RR-05-16},
  year={2005},
  doi={10.1002/j.2333-8504.2005.tb01993.x},
  url={https://www.ets.org/research/policy_research_reports/publications/report/2005/ibai.html}
}

@inproceedings{ma2022kscd,
  title={Knowledge-Sensed Cognitive Diagnosis for Intelligent Education Platforms},
  author={Ma, Haiping and Li, Manwei and Wu, Le and Zhang, Haifeng and Cao, Yunbo and Zhang, Xingyi and Zhao, Xuemin},
  booktitle={Proceedings of the 31st ACM International Conference on Information and Knowledge Management},
  pages={1451--1460},
  year={2022},
  doi={10.1145/3511808.3557372}
}

@inproceedings{zhou2021context,
  title={Modeling Context-aware Features for Cognitive Diagnosis in Student Learning},
  author={Zhou, Yuqiang and Liu, Qi and Wu, Jinze and Wang, Fei and Huang, Zhenya and Tong, Wei and Xiong, Hui and Chen, Enhong and Ma, Jianhui},
  booktitle={Proceedings of the 27th ACM SIGKDD Conference on Knowledge Discovery and Data Mining},
  pages={2420--2428},
  year={2021},
  doi={10.1145/3447548.3467264}
}

@inproceedings{tong2021irr,
  title={Item Response Ranking for Cognitive Diagnosis},
  author={Tong, Shiwei and Liu, Qi and Yu, Runlong and Huang, Wei and Huang, Zhenya and Pardos, Zachary A. and Jiang, Weijie},
  booktitle={Proceedings of the Thirtieth International Joint Conference on Artificial Intelligence},
  pages={1750--1756},
  year={2021},
  doi={10.24963/ijcai.2021/241}
}

@inproceedings{cheng2019dirt,
  title={{DIRT}: Deep Learning Enhanced Item Response Theory for Cognitive Diagnosis},
  author={Cheng, Song and Liu, Qi and Chen, Enhong and Huang, Zai and Huang, Zhenya and Chen, Yiying and Ma, Haiping and Hu, Guoping},
  booktitle={Proceedings of the 28th ACM International Conference on Information and Knowledge Management},
  pages={2397--2400},
  year={2019},
  doi={10.1145/3357384.3358070}
}

@article{song2023cmncd,
  title={A deep cross-modal neural cognitive diagnosis framework for modeling student performance},
  author={Song, Lingyun and He, Mengting and Shang, Xuequn and Yang, Chen and Liu, Jun and Yu, Mengzhen and Lu, Yu},
  journal={Expert Systems with Applications},
  volume={230},
  pages={120675},
  year={2023},
  doi={10.1016/j.eswa.2023.120675},
  publisher={Elsevier}
}

@article{zhang2025llm4cd,
  title={{LLM4CD}: Leveraging Large Language Models for Open-World Knowledge Augmented Cognitive Diagnosis},
  author={Zhang, Weiming and Fu, Lingyue and Li, Qingyao and Du, Kounianhua and Lin, Jianghao and Yu, Jingwei and Xia, Wei and Zhang, Weinan and Tang, Ruiming and Yu, Yong},
  journal={arXiv preprint arXiv:2505.13492},
  year={2025},
  url={https://arxiv.org/abs/2505.13492}
}

@article{ma2025lcst,
  title={Large Language Models Are Zero-Shot Cross-Domain Diagnosticians in Cognitive Diagnosis},
  author={Ma, Haiping and Wang, Changqian and Song, Siyu and Yang, Shangshang and Zhang, Limiao and Zhang, Xingyi},
  journal={Frontiers of Digital Education},
  volume={2},
  number={2},
  pages={17},
  year={2025},
  doi={10.1007/s44366-025-0054-y},
  url={https://journal.hep.com.cn/fde/EN/10.1007/s44366-025-0054-y}
}

@inproceedings{kochmar2024bertirt,
  title={{BERT-IRT}: Accelerating Item Piloting with {BERT} Embeddings and Explainable {IRT} Models},
  author={Yancey, Kevin P. and Runge, Andrew and LaFlair, Geoffrey and Mulcaire, Phoebe},
  booktitle={Proceedings of the 19th Workshop on Innovative Use of NLP for Building Educational Applications},
  pages={428--438},
  year={2024},
  address={Mexico City, Mexico},
  publisher={Association for Computational Linguistics},
  url={https://aclanthology.org/2024.bea-1.35/}
}

@inproceedings{veeramani2024llmitem,
  title={Large Language Model-based Pipeline for Item Difficulty and Response Time Estimation for Educational Assessments},
  author={Veeramani, Hariram and Thapa, Surendrabikram and Shankar, Natarajan Balaji and Alwan, Abeer},
  booktitle={Proceedings of the 19th Workshop on Innovative Use of NLP for Building Educational Applications},
  pages={561--566},
  year={2024},
  address={Mexico City, Mexico},
  publisher={Association for Computational Linguistics},
  url={https://aclanthology.org/2024.bea-1.49/}
}

@article{kasneci2023chatgpt,
  title={ChatGPT for good? On opportunities and challenges of large language models for education},
  author={Kasneci, Enkelejda and Sessler, Kathrin and Kuchemann, Stefan and Bannert, Maria and Dementieva, Daryna and Fischer, Frank and Gasser, Urs and Groh, Georg and Gunnemann, Stephan and Hullermeier, Eyke},
  journal={Learning and Individual Differences},
  volume={103},
  pages={102274},
  year={2023},
  doi={10.1016/j.lindif.2023.102274},
  publisher={Elsevier}
}

@misc{openai2024embedding3,
  title        = {New embedding models and API updates},
  author       = {{OpenAI}},
  year         = {2024},
  month        = jan,
  day          = {25},
  howpublished = {\url{https://openai.com/index/new-embedding-models-and-api-updates/}}
}

@misc{anthropic2025sonnet4,
  title={Introducing Claude 4: Sonnet 4 and Opus 4},
  author={Anthropic},
  year={2025},
  month={may},
  day={22},
  howpublished={\url{https://www.anthropic.com/news/claude-4}}
}
